\documentclass[preprint,12pt]{elsarticle}

\usepackage{amssymb}
\usepackage{amsmath,amssymb,amsfonts}
\usepackage{algorithmic}
\usepackage{algorithm}
\usepackage{graphicx}
\usepackage{textcomp}
\usepackage{verbatim}
\usepackage{multirow}
\usepackage{tabu}
\usepackage{makecell}
\usepackage{color}

\usepackage{wrapfig}
\usepackage{url}

\journal{Neurocomputing}

\begin{document}

\begin{frontmatter}



\title{On Hyperparameter Optimization of Machine Learning Algorithms: Theory and Practice}


\author{Li Yang and Abdallah Shami}
\ead{\{lyang339, abdallah.shami\}@uwo.ca}


\address{Department of Electrical and Computer Engineering, University of Western Ontario, 1151 Richmond St, London, Ontario, Canada N6A 3K7}

\begin{abstract}
Machine learning algorithms have been used widely in various applications and areas. To fit a machine learning model into different problems, its hyper-parameters must be tuned. Selecting the best hyper-parameter configuration for machine learning models has a direct impact on the model's performance. It often requires deep knowledge of machine learning algorithms and appropriate hyper-parameter optimization techniques. Although several automatic optimization techniques exist, they have different strengths and drawbacks when applied to different types of problems. In this paper, optimizing the hyper-parameters of common machine learning models is studied. We introduce several state-of-the-art optimization techniques and discuss how to apply them to machine learning algorithms. Many available libraries and frameworks developed for hyper-parameter optimization problems are provided, and some open challenges of hyper-parameter optimization research are also discussed in this paper. Moreover, experiments are conducted on benchmark datasets to compare the performance of different optimization methods and provide practical examples of hyper-parameter optimization. This survey paper will help industrial users, data analysts, and researchers to better develop machine learning models by identifying the proper hyper-parameter configurations effectively.\footnote{
General Hyperparameter Optimization Code and Tutorials: \url{https://github.com/LiYangHart/Hyperparameter-Optimization-of-Machine-Learning-Algorithms}}
\end{abstract}


\end{frontmatter}


\section{Introduction}
\label{sec:introduction}
Machine learning (ML) algorithms have been widely used in many applications domains, including advertising, recommendation systems, computer vision, natural language processing, and user behavior analytics \cite{MLsc}. This is because they are generic and demonstrate high performance in data analytics problems. Different ML algorithms are suitable for different types of problems or datasets \cite{SOAML}. In general, building an effective machine learning model is a complex and time-consuming process that involves determining the appropriate algorithm and obtaining an optimal model architecture by tuning its hyper-parameters (HPs) \cite{AMLSC}. 

Two types of parameters exist in machine learning models: one that can be initialized and updated through the data learning process (\textit{e.g.}, the weights of neurons in neural networks), named model parameters; while the other, named hyper-parameters, cannot be directly estimated from data learning and must be set before training a ML model because they define the architecture of a ML model \cite{2Ps}. Hyper-parameters are the parameters that are used to either configure a ML model (\textit{e.g.}, the penalty parameter $C$ in a support vector machine, and the learning rate to train a neural network) or to specify the algorithm used to minimize the loss function (\textit{e.g.}, the activation function and optimizer types in a neural network, and the kernel type in a support vector machine) \cite{parameters}.

To build an optimal ML model, a range of possibilities must be explored. The process of designing the ideal model architecture with an optimal hyper-parameter configuration is named hyper-parameter tuning. Tuning hyper-parameters is considered a key component of building an effective ML model, especially for tree-based ML models and deep neural networks, which have many hyper-parameters \cite{AMLB}. Hyper-parameter tuning process is different among different ML algorithms due to their different types of hyper-parameters, including categorical, discrete, and continuous hyper-parameters \cite{EHPO}. Manual testing is a traditional way to tune hyper-parameters and is still prevalent in graduate student research, although it requires a deep understanding of the used ML algorithms and their hyper-parameter value settings \cite{ADL}. However, manual tuning is ineffective for many problems due to certain factors, including a large number of hyper-parameters, complex models, time-consuming model evaluations, and non-linear hyper-parameter interactions. These factors have inspired increased research in techniques for automatic optimization of hyper-parameters; so-called hyper-parameter optimization (HPO) \cite{BBHPO}. The main aim of HPO is to automate hyper-parameter tuning process and make it possible for users to apply machine learning models to practical problems effectively \cite{AMLSC}. The optimal model architecture of a ML model is expected to be obtained after a HPO process. Some important reasons for applying HPO techniques to ML models are as follows \cite{AMLB}:  

\begin{enumerate}
\item It reduces the human effort required, since many ML developers spend considerable time tuning the hyper-parameters, especially for large datasets or complex ML algorithms with a large number of hyper-parameters.

\item It improves the performance of ML models. Many ML hyper-parameters have different optimums to achieve best performance in different datasets or problems.

\item It makes the models and research more reproducible. Only when the same level of hyper-parameter tuning process is implemented can different ML algorithms be compared fairly; hence, using a same HPO method on different ML algorithms also helps to determine the most suitable ML model for a specific problem.
\end{enumerate}

It is crucial to select an appropriate optimization technique to detect optimal hyper-parameters. Traditional optimization techniques may be unsuitable for HPO problems, since many HPO problems are non-convex or non-differentiable optimization problems, and may result in a local instead of a global optimum \cite{ASHPO}. Gradient descent-based methods are a common type of traditional optimization algorithm that can be used to tune continuous hyper-parameters by calculating their gradients \cite{GBad}. For example, the learning rate in a neural network can be optimized by a gradient-based method.

Compared with traditional optimization methods like gradient descent, many other optimization techniques are more suitable for HPO problems, including decision-theoretic approaches, Bayesian optimization models, multi-fidelity optimization techniques, and metaheuristics algorithms \cite{EHPO}. Apart from detecting continuous hyper-parameters, many of these algorithms also have the capacity to effectively identify discrete, categorical, and conditional hyper-parameters. 

Decision-theoretic methods are based on the concept of defining a hyper-parameter search space and then detecting the hyper-parameter combinations in the search space, ultimately selecting the best-performing hyper-parameter combination. Grid search (GS) \cite{AHPO} is a decision-theoretic approach that involves exhaustively searching for a fixed domain of hyper-parameter values. Random search (RS) \cite{RS} is another decision-theoretic method that randomly selects hyper-parameter combinations in the search space, given limited execution time and resources. In GS and RS, each hyper-parameter configuration is treated independently.

Unlike GS and RS, Bayesian optimization (BO) \cite{BOHP} models determine the next hyper-parameter value based on the previous results of tested hyper-parameter values, which avoids many unnecessary evaluations; thus, BO can detect the optimal hyper-parameter combination within fewer iterations than GS and RS. To be applied to different problems, BO can model the distribution of the objective function using different models as the surrogate function, including Gaussian process (GP), random forest (RF), and tree-structured Parzen estimators (TPE) models \cite{surrogates}. BO-RF and BO-TPE can retain the conditionality of variables \cite{surrogates}. Thus, they can be used to optimize conditional hyper-parameters, like the kernel type and the penalty parameter $C$ in a support vector machine (SVM). However, since BO models work sequentially to balance the exploration of unexplored areas and the exploitation of currently-tested regions, it is difficult to parallelize them. 

Training a ML model often takes considerable time and space. Multi-fidelity optimization algorithms are developed to tackle problems with limited resources, and the most common ones being bandit-based algorithms. Hyperband \cite{Hyperband} is a popular bandit-based optimization technique that can be considered an improved version of RS. It generates small versions of datasets and allocates a same budget to each hyper-parameter combination. In each iteration of Hyperband, poorly-performing hyper-parameter configurations are eliminated to save time and resources. 

Metaheuristic algorithms are a set of techniques used to solve complex, large search space and non-convex optimization problems to which HPO problems belong \cite{HumanAML}. Among all metaheuristic methods, genetic algorithm (GA) \cite{GA2} and particle swarm optimization (PSO) \cite{PSODL} are the two most prevalent metaheuristic algorithms used for HPO problems. Genetic algorithms detect well-performing hyper-parameter combinations in each generation, and pass them to the next generation until the best-performing combination is identified. In PSO algorithms, each particle communicates with other particles to detect and update the current global optimum in each iteration until the final optimum is detected. Metaheuristics can efficiently explore the search space to detect optimal or near-optimal solutions. Hence, they are particularly suitable for the HPO problems with large configuration space due to their high efficiency. For instance, they can be used in deep neural networks (DNNs) which have a large configuration space with multiple hyper-parameters, including the activation and optimizer types, the learning rate, drop-out rate, etc.

Although using HPO algorithms to tune the hyper-parameters of ML models greatly improves the model performance, certain other aspects, like their computational complexity, still have much room for improvement. On the other hand, since different HPO models have their own advantages and suitable problems, overviewing them is necessary for proper optimization algorithm selection in terms of different types of ML models and problems. 

This paper makes the following contributions: 
\begin{enumerate}
\item It reviews common ML algorithms and their important hyper-parameters. 
\item It analyzes common HPO techniques, including their benefits and drawbacks, to help apply them to different ML models by appropriate algorithm selection in practical problems. 
\item It surveys common HPO libraries and frameworks for practical use. 
\item It discusses the open challenges and research directions of the HPO research domain. 
\end{enumerate}

In this survey paper, we begin with a comprehensive introduction of the common optimization techniques used in ML hyper-parameter tuning problems. Section 2 introduces the main concepts of mathematical optimization and hyper-parameter optimization, as well as the general HPO process. In Section 3, we discuss the key hyper-parameters of common ML models that need to be tuned. Section 4 covers the various state-of-the-art optimization approaches that have been proposed for tackling HPO problems. In Section 5, we analyze different HPO methods and discuss how they can be applied to ML algorithms. In Section 6, we provide an introduction to various public libraries and frameworks that are developed to implement HPO. Section 7 presents and discusses the experimental results of using HPO on benchmark datasets for HPO method comparison and practical use case demonstration. In Section 8, we discuss several research directions and open challenges that should be considered to improve current HPO models or develop new HPO approaches. We conclude the paper in Section 9.
\section{Mathematical Optimization and Hyper-parameter Optimization Problems}
The key process of machine learning is to solve optimization problems. To build a ML model, its weight parameters are initialized and optimized by an optimization method until the objective function approaches a minimum value or the accuracy approaches a maximum value \cite{mathopt1}. Similarly, hyper-parameter optimization methods aim to optimize the architecture of a ML model by detecting the optimal hyper-parameter configurations. In this section, the main concepts of mathematical optimization and hyper-parameter optimization for machine learning models are discussed.
\subsection{Mathematical Optimization}
Mathematical optimization is the process of finding the best solution from a set of available candidates to maximize or minimize the objective function \cite{mathopt1}. Generally, optimization problems can be classified as constrained or unconstrained optimization problems based on whether they have constraints for the decision variables or the solution variables. 

In unconstrained optimization problems, a decision variable, $x$, can take any values from the one-dimensional space of all real numbers, $\mathbb{R}$. An unconstrained optimization problem can be denoted by \cite{mathopt2}:
\begin{equation}\min_{x \in \mathbb{R}} f(x),\end{equation}
where $f(x)$ is the objective function. 

On the other hand, most real-life optimization problems are constrained optimization problems. The decision variable $x$ for constrained optimization problems should be subject to certain constraints which could be mathematical equalities or inequalities. Therefore, constrained optimization problems or general optimization problems can be expressed as \cite{mathopt2}:
\begin{equation}
\begin{aligned}
&\min_{x} f(x) \\
&\text{subject to}\\
&\; g_{i}(x) \leq 0, i=1,2,\cdots, m,\\
&\; h_{j}(x) = 0, j=1,2,\cdots, p,\\
&x \in X,
\end{aligned}
\end{equation}
where $g_{i}(x),i=1,2,\cdots, m,$ are the inequality constraint functions; $h_{j}(x), j=1,2,\cdots, p,$ are the equality constraint function; and $X$ is the domain of $x$. 

The role of constraints is to limit the possible values of the optimal solution to certain areas of the search space, named the feasible region \cite{mathopt2}. Thus, the feasible region $D$ of $x$ can be represented by:
\begin{equation}
D=\{x \in X|g_{i}(x) \leq 0,h_{j}(x) = 0\}.
\end{equation}

To conclude, an optimization problem consists of three major components: a set of decision variables $x$, an objective function $f(x)$ to be either minimized or maximized, and a set of constraints that allow the variables to take on values in certain ranges (if it is a constrained optimization problem). Therefore, the goal of optimization tasks is to obtain the set of variable values that minimize or maximize the objective function while satisfying any applicable constraints. 

Regarding ML models, many HPO problems have certain constraints, like the feasible domain of the number of clusters in k-means, as well as time and space constraints. Therefore, constrained optimization techniques are widely-used in HPO problems \cite{AMLSC}.

For optimization problems, in many cases, only a local instead of a global optimum can be obtained. For example, to obtain the minimum of a problem, assuming $D$ is the feasible region of a decision variable $x$, a global minimum is the point $x^* \in D$ satisfying $f(x^*) \leq f(x) \: \forall x \in D$ , while a local minimum is a point $x^* \in D$ in a neighborhood $N$ satisfying $f\left(x^{*}\right) \leq f(x) \: \forall x \in N \cap D$ \cite{mathopt2} . Thus, the local optimum may only be an optimum in a small range instead of being the optimal solution in the entire feasible region.

A local optimum is only guaranteed to be the global optimum in convex functions \cite{convex}. Convex functions are the functions that only have one optimum. Therefore, continuing to search along the direction in which the objective function decreases can detect the global minimum value. A function $f(x)$ is a convex function if for $\forall x_{1}, x_{2} \in X, \forall t \in[0,1]$, 
\begin{equation} \quad f\left(t x_{1}+(1-t) x_{2}\right) \leq t f\left(x_{1}\right)+(1-t) f\left(x_{2}\right),\end{equation}
where $X$ is the domain of decision variables, and $t$ is a coefficient in the range of [0,1].

An optimization problem is a convex optimization problem only when the objective function $f(x)$ is a convex function and the feasible region $C$ is a convex set, denoted by \cite{convex}:
\begin{equation}
\begin{aligned}
&\min_{x} f(x) \\
&\text{subject to} \;\; x \in C.
\end{aligned}
\end{equation}

On the other hand, nonconvex functions have multiple local optimums, but only one of these optimums is the global optimum. Most ML and HPO problems are nonconvex optimization problems. Thus, utilizing inappropriate optimization methods may only result in a local instead of a global optimum. 

There are many traditional methods that can be used to solve optimization problems, including gradient descent, Newton’s method, conjugate gradient, and heuristic optimization methods \cite{mathopt1}. Gradient descent is a commonly-used optimization method that uses the negative gradient direction as the search direction to move towards the optimum. However, gradient descent cannot guarantee to detect the global optimum unless the objective function is a convex function. Newton’s method uses the inverse matrix of the Hessian matrix to obtain the optimum. Newton’s method has faster convergence speed than gradient descent, but often requires more time and larger space than gradient descent to store and calculate the Hessian matrix. Conjugate gradient searches along the conjugated directions constructed by the gradient of known data points to detect the optimum. Conjugate gradient has faster convergence speed than gradient descent, but its calculation of conjugate gradient is more complex. Unlike other traditional methods, heuristic methods use empirical rules to solve the optimization problems instead of following systematical steps to obtain the solution. Heuristic methods can often detect the approximate global optimum within a few iterations, but cannot guarantee to detect the global optimum \cite{mathopt1}.
\subsection{Hyper-parameter Optimization Problem Statement}

During the design process of ML models, effectively searching the hyper-parameters' space using optimization techniques can identify the optimal hyper-parameters for the models. The hyper-parameter optimization process consists of four main components: an estimator (a regressor or a classifier) with its objective function, a search space (configuration space), a search or optimization method used to find hyper-parameter combinations, and an evaluation function to compare the performance of different hyper-parameter configurations.  

The domain of a hyper-parameter can be continuous (\textit{e.g.}, learning rate), discrete (\textit{e.g.}, number of clusters), binary (\textit{e.g.}, whether to use early stopping or not), or categorical (\textit{e.g.}, type of optimizer). Therefore, hyper-parameters are classified as continuous, discrete, and categorical hyper-parameters. For continuous and discrete hyper-parameters, their domains are usually bounded in practical applications \cite{AHPO} \cite{UBO}. On the other hand, the hyper-parameter configuration space sometimes contains conditionality. A hyper-parameter may need to be used or tuned depending on the value of another hyper-parameter, called a conditional hyper-parameter \cite{ASHPO}. For instance, in SVM, the degree of the polynomial kernel function only needs to be tuned when the kernel type is chosen to be polynomial.

In simple cases, all hyper-parameters can take unrestricted real values, and the feasible set $X$ of hyper-parameters can be a real-valued $n$-dimensional vector space. However, in most cases, the hyper-parameters of a ML model often take on values from different domains and have different constraints, so their optimization problems are often complex constrained optimization problems \cite{HPOpro}. For instance, the number of considered features in a decision tree should be in the range of 0 to the number of features, and the number of clusters in k-means should not be larger than the size of data points. Additionally, categorical features can often only take several certain values, like the limited choices of the activation function and the optimizer of a neural network. Therefore, the feasible domain of $X$ often has a complex structure, which increases the problems' complexity \cite{HPOpro}. 

In general, for a hyper-parameter optimization problem, the aim is to obtain \cite{PSODL}: 
\begin{equation}
x^{*}=\arg \min _{x \in X} f(x),
\end{equation}
where $f(x)$ is the objective function to be minimized, such as the error rate or the root mean squared error (RMSE); $x^{*}$ is the hyper-parameter configuration that produces the optimum value of $f(x)$; and a hyper-parameter $x$ can take any value in the search space $X$. 

The aim of HPO is to achieve optimal or near-optimal model performance by tuning hyper-parameters within the given budgets \cite{AMLSC}. The mathematical expression of the function $f$ varies, depending on the objective function of the chosen ML algorithm and the performance metric function. Model performance can be evaluated by various metrics, like accuracy, RMSE, F1-score, and false alarm rate. On the other hand, in practice, time budgets are an essential constraint for optimizing HPO models and must be considered. It often requires a massive amount of time to optimize the objective function of a ML model with a reasonable number of hyper-parameter configurations. Every time a hyper-parameter value is tested, the entire ML model needs to be retrained, and the validation set needs to be processed to generate a score that reflects the model performance.

The main process of HPO is as follows \cite{ASHPO}:
\begin{enumerate}
\item Select the objective function and the performance metrics;
\item Select the hyper-parameters that require tuning, summarize their types, and determine the appropriate optimization technique;
\item Train the ML model using the default hyper-parameter configuration or common values as the baseline model;
\item Start the optimization process with a large search space as the hyper-parameter feasible domain determined by manual testing and/or domain knowledge;
\item Narrow the search space based on the regions of currently-tested well-performing hyper-parameter values, or explore new search spaces if necessary.
\item Return the best-performing hyper-parameter configuration as the final solution. 
\end{enumerate}

However, most traditional optimization techniques \cite{OML} are unsuitable for HPO, since HPO problems are different from traditional optimization problems in the following aspects \cite{ASHPO}:
\begin{enumerate}
\item The optimization target, the objective function of ML models, is usually a non-convex and non-differentiable function. Therefore, many traditional optimization methods designed to solve convex or differentiable optimization problems are often unsuitable for HPO problems, since these methods may return a local optimum instead of a global optimum. Additionally, an optimization target lacking smoothness makes certain traditional derivative-free optimization models perform poorly for HPO problems \cite{AUTOMS}.

\item The hyper-parameters of ML models include continuous, discrete, categorical, and conditional hyper-parameters. Thus, many traditional numerical optimization methods \cite{NumericalO} that only aim to tackle numerical or continuous variables are unsuitable for HPO problems.

\item It is often computationally expensive to train a ML model on a large-scale dataset. HPO techniques sometimes use data sampling to obtain approximate values of the objective function. Thus, effective optimization techniques for HPO problems should be able to use these approximate values. However, function evaluation time is often ignored in many black-box optimization (BBO) models, so they often require exact instead of approximate objective function values. Consequently, many BBO algorithms are often unsuitable for HPO problems with limited time and resource budgets. 
\end{enumerate}

Therefore, appropriate optimization algorithms should be applied to HPO problems to identify optimal hyper-parameter configurations for ML models.

\section{Hyper-parameters in Machine Learning Models}
To boost ML models by HPO, firstly, we need to find out what the key hyper-parameters are that people need to tune to fit the ML models into specific problems or datasets.

In general, ML models can be classified as supervised and unsupervised learning algorithms, based on whether they are built to model labeled or unlabeled datasets \cite{superAM}. Supervised learning algorithms are a set of machine learning algorithms that map input features to a target by training on labeled data, and mainly include linear models, k-nearest neighbors (KNN), support vector machines (SVM), naïve Bayes (NB), decision-tree-based models, and deep learning (DL) algorithms \cite{supervised}. Unsupervised learning algorithms are used to find patterns from unlabeled data and can be divided into clustering and dimensionality reduction algorithms based on their aims. Clustering methods mainly include k-means, density-based spatial clustering of applications with noise (DBSCAN), hierarchical clustering, and expectation-maximization (EM); while two common dimensionality reduction algorithms are principal component analysis (PCA) and linear discriminant analysis (LDA) \cite{sklearnbook}.  Moreover, there are several ensemble learning methods that combine different singular models to further improve model performance, like voting, bagging, and AdaBoost. In this paper, the important hyper-parameters of common ML models are studied based on their names in Python libraries, including scikit-learn (sklearn) \cite{sklearn}, XGBoost \cite{xgboost}, and Keras\cite{keras}.

\subsection{Supervised Learning Algorithms}
In supervised learning, both the input $x$ and the output $y$ are available, and the goal is to obtain an optimal predictive model function $f^*$ to minimize the cost function $\mathcal{L}(f(x),y)$ that models the error between the estimated output and ground-truth labels. The predictive model function $f$ varies based on its model structure. With limited model architectures determined by different hyper-parameter configurations, the domain of the ML model function $f$ is restricted to a set of functions $F$. Thus, the optimal predictive model $f^*$ can be obtained by \cite{SL}:
\begin{equation}
f^{*}=\arg \min _{f \in F} \frac{1}{n} \sum_{i=1}^{n} \mathcal{L}\left(f\left(x_{i}\right), y_{i}\right)
\end{equation}
where $n$ is the number of training data points, $x_i$ is the feature vector of the $i$-th instance, $y_i$ is the corresponding actual output, and $L$ is the cost function value of each sample. 

Many different loss functions exist in supervised learning algorithms, including the square of Euclidean distance, cross-entropy, information gain, etc. \cite{SL}. On the other hand, different ML algorithms generate different predictive model architectures based on different hyper-parameter configurations, which will be discussed in detail in this subsection.

\subsubsection{Linear Models}
In general, supervised learning models can be classified as regression and classification techniques when used to predict continuous or discrete target variables, respectively. Linear regression \cite{ML} is a typical regression model that predicts a target $y$ by the following equation:
\begin{equation}
\hat{y}(\mathbf{w}, \mathbf{x})=w_{0}+w_{1} x_{1}+\ldots+w_{p} x_{p},
\end{equation}
where the target variable $y$ is expected to be a linear combination of $p$ input features $\mathbf{x}=\left(x_{1}, \cdots x_{p}\right)$, and $\hat{y}$ is the predicted value. The weight vector $\mathbf{w}=\left(w_{1}, \cdots w_{p}\right)$ is designated as an attribute 'coef\_', and $w_{0}$ is defined as another attribute 'intercept\_' in the linear model of sklearn. Usually, no hyper-parameter needs to be tuned in linear regression. A linear model's performance mainly depends on how well the problem or data follows a linear distribution. 

To improve the original linear regression models, ridge regression was proposed in \cite{ridge}. Ridge regression imposes a penalty on the coefficients, and aims to minimize the objective function \cite{ridgelasso}:  
\begin{equation}
\alpha\|w\|_{2}^{2}+\sum_{i=1}^{p}\left(y_{i}-w_{i} \cdot x_{i}\right)^{2},
\end{equation}
where $\|w\|_{2}$ is the $L_2$-norm of the coefficient vector, and $\alpha$ is the regularization strength. A larger value of $\alpha$ indicates a larger amount of shrinkage; thus, the coefficients are also more robust to collinearity.

Lasso regression \cite{lasso} is another linear model used to estimate sparse coefficients, consisting of a linear model with an $L_1$ priori added regularization term. It aims to minimize the objective function \cite{ridgelasso}:
\begin{equation}
\alpha\|w\|_{1}+\sum_{i=1}^{p}\left(y_{i}-w_{i} \cdot x_{i}\right)^{2},
\end{equation}
where $\alpha$ is the regularization strength and $\|w\|_{1}$ is the $L_1$-norm of the coefficient vector. Therefore, the regularization strength $\alpha$ is an crucial hyper-parameter of both ridge and lasso regression models. 

Logistic regression (LR) \cite{LoR} is a linear model used for classification problems. In LR, its cost function may be different, depending on the regularization method chosen for the penalization. There are three main types of regularization methods in LR: $L_1$-norm,  $L_2$-norm, and elastic-net regularization \cite{LRnorms}.

Therefore, the first hyper-parameter that needs to be tuned in LR is to the regularization method used in the penalization, 'l1', 'l2', 'elasticnet' or 'none', which is called 'penalty' in sklearn. The coefficient, '$C$', is another essential hyper-parameter that determines the regularization strength of the model. In addition, the 'solver' type, representing the optimization algorithm type, can be set to 'newton-cg', 'lbfgs', 'liblinear', 'sag', or 'saga' in LR. The 'solver' type has correlations with 'penalty' and '$C$', so they are conditional hyper-parameters. 

\subsubsection{KNN}
K-nearest neighbor (KNN) is a simple ML algorithm that is used to classify data points by calculating the distances between different data points. In KNN, the predicted class of each test sample is set to the class to which most of its k-nearest neighbors in the training set belong. 

Assuming the training set $T=\{(x_1,y_1),(x_2,y_2),\cdots,(x_n,y_n)\}$, $x_i$ is the feature vector of an instance, and $y_i\in\{c_1,c_2,\cdots,c_m\}$ is the class of the instance, $i=\left(1,2, \cdots n\right)$, for a test instance $x$, its class $y$ can be denoted by \cite{KNN}:
\begin{equation}
y=\arg \max _{c_j}\sum_{x_i \in N_k(x)} I\left(y_{i} = c_{j}\right), i=1,2,\cdots,n; j=1,2,\cdots,m,
\end{equation}
where $I(x)$ is an indicator function, $I=1$ when $y_i=c_j$, otherwise $I=0$; $N_k(x)$ is the field involving the k-nearest neighbors of $x$. 

In KNN, the number of considered nearest neighbors, $k$, is the most crucial hyper-parameter \cite{kinknn}. If $k$ is too small, the model will be under-fitting; if $k$ is too large, the model will be over-fitting and require high computational time. In addition, the weighted function used in the prediction can also be chosen from 'uniform' (points are weighted equally) or 'distance' (points are weighted by the inverse of their distance), depending on specific problems. The distance metric and the power parameter of the Minkowski metric can also be tuned as it can result in minor improvement. Lastly, the 'algorithm' used to compute the nearest neighbors can also be chosen from a ball tree, a k-dimensional (KD) tree, or a brute force search. Typically, the model can determine the most appropriate algorithm itself by setting the 'algorithm' to 'auto' in sklearn \cite{sklearn}.
\subsubsection{SVM}
A support vector machines (SVM) \cite{SVM1} is a supervised learning algorithm that can be used for both classification and regression problems. SVM algorithms are based on the concept of mapping data points from low-dimensional into high-dimensional space to make them linearly separable; a hyperplane is then generated as the classification boundary to partition data points \cite{SVMme}. Assuming there are $n$ data points, the objective function of SVM is \cite{SVMme2}: 
\begin{equation}
\arg \min _{\mathbf{w}}\left\{\frac{1}{n} \sum_{i=1}^{n} \max \left\{0,1-y_{i}f\left(x_{i}\right)\right\}+C \mathbf{w}^{T} \mathbf{w}\right\},
\end{equation}
where $\mathbf{w}$ is a normalization vector; $C$ is the penalty parameter of the error term, which is an important hyper-parameter of all SVM models.  

The kernel function $f(x)$, which is used to measure the similarity between two data points $x_i$ and $x_j$, can be chosen from multiple types of kernels in SVM models. Therefore, the kernel type would be a vital hyper-parameter to be tuned. Common kernel types in SVM include linear kernels, radial basis function (RBF), polynomial kernels, and sigmoid kernels. 

The different kernel functions can be denoted as follows \cite{SVMkernel}: 
\begin{enumerate}
\item Linear kernel:
\begin{equation}
f(x)=x_{i}^{T} x_{j};
\end{equation}

\item Polynomial kernel:
\begin{equation}
f(x)=\left(\gamma x_{i}^{T} x_{j}+r\right)^{d};
\end{equation}

\item RBF kernel:
\begin{equation}
f(x)=\exp \left(-\gamma\left\|x-x^{\prime}\right\|^{2}\right);
\end{equation}

\item Sigmoid kernel: 
\begin{equation}
f(x)=\left(\tanh \left(\gamma x_{i}^{T} x_{j}+r\right)\right);
\end{equation}
\end{enumerate}

As shown in the kernel function equations, a few other different hyper-parameters need to be tuned after a kernel type is chosen. The coefficient $\gamma$, denoted by 'gamma' in sklearn, is the conditional hyper-parameter of the 'kernel type' hyper-parameter when it is set to polynomial, RBF, or sigmoid; $r$, specified by 'coef0' in sklearn, is the conditional hyper-parameter of polynomial and sigmoid kernels. Moreover, the polynomial kernel has an additional conditional hyper-parameter $d$ representing the 'degree' of the polynomial kernel function. In support vector regression (SVR) models, there is another hyper-parameter, 'epsilon', indicating the distance error to of its loss function \cite{sklearn}.

\subsubsection{Naïve Bayes}
Naïve Bayes (NB) \cite{NB1} algorithms are supervised learning algorithms based on Bayes' theorem. 
Assuming there are $n$ dependent features $x_{1}, \cdots x_{n}$ and a target variable $y$, the objective function of naïve Bayes can be denoted by:
\begin{equation}
\hat{y}=\arg \max _{y} P(y) \prod_{i=1}^{n} P\left(x_{i} | y\right),
\end{equation}
where $P(y)$ is the probability of a value $y$, and $P\left(x_{i} | y\right)$ is the posterior probabilities of $x_{i}$ given the values of $y$.
Regarding the different assumptions of the distribution of $P\left(x_{i} | y\right)$, there are different types of naïve Bayes classifiers. The four main types of NB models are: Bernoulli NB, Gaussian NB, multinomial NB, and complement NB \cite{NB2}. 

For Gaussian NB \cite{GNB}, the likelihood of features is assumed to follow a Gaussian distribution:  
\begin{equation}
P\left(x_{i} | y\right)=\frac{1}{\sqrt{2 \pi \sigma_{y}^{2}}} \exp \left(-\frac{\left(x_{i}-\mu_{y}\right)^{2}}{2 \sigma_{y}^{2}}\right).
\end{equation}

The maximum likelihood method is used to calculate the mean value, $\mu_{y}$, and the variance, $\sigma_{y}^2$. Normally, there is not any hyper-parameter that needs to be tuned for Gaussian NB. The performance of a Gaussian NB model mainly depends on how well the dataset follows  Gaussian distributions.

Multinomial NB \cite{MNB} is designed for multinomially-distributed data based on the naïve Bayes algorithm. Assuming there are $n$ features, and $\theta_{y i}$ is the distribution of each value of the target variable $y$, which equals the conditional probability $P\left(x_{i} | y\right)$ when a feature value $i$ is involved in a data point belonging to the class $y$. Based on the concept of relative frequency counting, $\theta_{y}$ can be estimated by a smoothed version of $\theta_{y i}$ \cite{sklearn}:  
\begin{equation}
\hat{\theta}_{y i}=\frac{N_{y i}+\alpha}{N_{y}+\alpha n},
\end{equation}
where $N_{y i}$ is the number of times when feature $i$ is in a data point belonging to class $y$, and $N_{y}$ is the sum of all $N_{y i}$ ($i=0, 1, 2,  \cdots, n$).
The smoothing priors $\alpha \geq 0$ are used for features that are not in the learning samples. When $\alpha=1$, it is called Laplace smoothing; when $\alpha<1$, it is called Lidstone smoothing. 

Complement NB \cite{CNB} is an improved version of the standard multinomial NB algorithm and is suitable for processing imbalanced data, while Bernoulli NB \cite{BNB} requires samples to have binary-valued feature vectors so that the data can follow multivariate Bernoulli distributions. They both have the additive (Laplace/Lidstone) smoothing parameter, $\alpha$, as the main hyper-parameter that needs tuning. To conclude, for naïve Bayes algorithms, developers often do not need to tune hyper-parameters or only need to tune the smoothing parameter $\alpha$, which is a continuous hyper-parameter.
\subsubsection{Tree-based Models}
Decision tree (DT) \cite{DT} is a common classification method that uses a tree-structure to model decisions and possible consequences by summarizing a set of classification rules from the data. A DT has three main components: a root node representing the entire data; multiple decision nodes indicating decision tests and sub-node splits over each feature; and several leaf nodes representing the result classes \cite{n3}. DT algorithms recursively split the training set with better feature values to achieve good decisions on each subset. Pruning, which means removing some of the sub-nodes of decision nodes, is used in DT to avoid over-fitting. Since a deeper tree has more sub-trees to make more accurate decisions, the maximum tree depth, 'max\_depth', is an essential hyper-parameter that controls the complexity of DT algorithms \cite{IDSme}.

There are many other important HPs to be tuned to build effective DT models \cite{DTHPsk}. Firstly, the quality of splits can be measured by setting a measuring function, denoted by 'criterion' in sklearn. Gini impurity and information gain are the two main types of measuring functions. The split selection method, 'splitter', can also be set to 'best' to choose the best split, or 'random' to select a random split. The number of considered features to generate the best split, 'max\_features', can also be tuned as a feature selection process. Moreover, there are several discrete hyper-parameters related to the splitting process: the minimum number of data points to split a decision node or to obtain a leaf node, denoted by 'min\_samples\_split' and 'min\_samples\_leaf', respectively; the 'max\_leaf\_nodes', indicating the maximum number of leaf nodes, and the 'min\_weight\_fraction\_leaf' that means the minimum weighted fraction of the total weights, can also be tuned to improve model performance \cite{sklearn} \cite{DTHPsk}. 

Based on the concept of DT models, many decision-tree-based ensemble algorithms have been proposed to improve model performance by combining multiple decision trees, including random forest (RF), extra trees (ET), and extreme gradient boosting (XGBoost) models. RF \cite{RF} is an ensemble learning method that uses the bagging method to combine multiple decision trees. In RF, basic DTs are built on many randomly-generated subsets, and the class with the majority voting will be selected to be the final classification result \cite{RFour}. ET \cite{ET} is another tree-based ensemble learning method that is similar to RF, but it uses all samples to build DTs and randomly selects the feature sets. In addition, RF optimizes splits on DTs while ET randomly makes the splits. XGBoost \cite{xgboost} is a popular tree-based ensemble model designed for speed and performance improvement, which uses the boosting and gradient descent methods to combine basic DTs. In XGBoost,	the next input sample of a new DT will be related to the results of previous DTs. XGBoost aims to minimize the following objective function \cite{IDSme}:
\begin{equation}
Obj = -\frac{1}{2}\sum_{j=1}^t\frac{G_{j}^2}{H_{j}+\lambda}+\gamma t,
\end{equation}
where $t$ is the number of leaves in a decision tree, $G$ and $H$ are the sums of the first and second order gradient statistics of the cost function, $\gamma$ and $\lambda$ are the penalty coefficients.

Since tree-based ensemble models are built with decision trees as base learners, they have the same hyper-parameters as DT models, as described in this subsection. Apart from these hyper-parameters, RF, ET, and XGBoost all have another crucial hyper-parameter to be tuned, which is the number of decision trees to be combined, denoted by 'n\_estimators' in sklearn. XGBoost has several additional hyper-parameters, including \cite{XGHP}: 'min\_child\_weight' which means the minimum sum of weights in a child node; 'subsample' and 'colsample\_bytree' used to control the subsampling ratio of instances and features, respectively; and four continuous hyper-parameters — 'gamma', 'alpha', 'lambda', and 'learning\_rate' — indicating the minimum loss reduction for a split, $L_1$, and $L_2$ regularization term on weights, and the learning rate, respectively. 

\subsubsection{Ensemble Learning Algorithms}
Apart from tree-based ensemble models, there are several other generic ensemble learning methods that combine multiple singular ML models to achieve better model performance than any singular algorithms alone. Three common ensemble learning models — voting, bagging, and AdaBoost — are introduced in this subsection \cite{Ensemble}. 

Voting \cite{Ensemble} is a basic ensemble learning algorithm that uses the majority voting rule to combine singular estimators and generate a comprehensive estimator with improved accuracy. In sklearn, the voting method can be set to be 'hard' or 'soft', indicating whether to use majority voting or averaged predicted probabilities to determine the classification result. The list of selected single ML estimators and their weights can also be tuned in certain cases. For instance, a higher weight can be assigned to a better-performing singular ML model in a voting model.

Bootstrap aggregating \cite{Ensemble}, also named bagging, trains multiple base estimators on different randomly-extracted subsets to construct a final predictor \cite{bagging}. When using bagging methods, the first consideration should be the type and number of base estimators in the ensemble, denoted by 'base\_estimator' and 'n\_estimators', respectively. Then, the 'max\_samples' and 'max\_features', indicating the sample size and feature size to generate different subsets, can also be tuned.

AdaBoost \cite{Ensemble}, short for adaptive boosting, is an ensemble learning method that trains multiple base learners consecutively (weak learners), and later learners emphasize the mis-classified samples of previous learners; ultimately, a final strong learner is obtained. During this process, incorrectly-classified instances are retrained with other new instances, and their weights are adjusted so that the subsequent classifiers focus more on difficult cases, thereby gradually building a stronger classifier. In AdaBoost, the type of base estimator, 'base\_estimator', can be set to a decision tree or other methods. In addition, the maximum number of estimators at which boosting is terminated, 'n\_estimators', and the learning rate that shrinks the contribution of each classifier, should also be tuned to achieve a trade-off between these two hyper-parameters. 
\subsubsection{Deep Learning Models}
Deep learning (DL) algorithms are widely applied to various areas — like computer vision, natural language processing, and machine translation — since they have had great success solving many types of problems. DL models are based on the theory of artificial neural networks (ANNs). Common types of DL architectures include deep neural networks (DNNs), feedforward neural networks (FFNNs), deep belief networks (DBNs), convolutional neural networks (CNNs), recurrent neural networks (RNNs) and many more \cite{DL1}. All these DL models have similar hyper-parameters since they have similar underlying neural network architecture. Compared with other ML models, DL models benefit more from HPO since they often have many hyper-parameters that require tuning. 

The first set of hyper-parameters is related to the construction of a DL model; hence, named model design hyper-parameters. Since all neural network models have an input layer and an output layer, the complexity of a deep learning model mainly depends on the number of hidden layers and the number of neurons in each layer, which are two main hyper-parameters to build DL models \cite{DL2}. These two hyper-parameters are set and tuned according to the complexity of the datasets or the problems. DL models need to have enough capacity to model objective functions (or prediction tasks) while avoiding over-fitting. At the next stage, certain function types need to be set or tuned. The first function type to configure is the loss function type, which is chosen mainly based on the problem type (\textit{e.g.}, binary cross-entropy for binary classification problems, multi-class cross-entropy for multi-classification problems, and RMSE for regression problems). Another important hyper-parameter is the activation function type used to model non-linear functions, which be set to 'softmax', 'rectified linear unit (ReLU)', 'sigmoid', 'tanh', or 'softsign'. Lastly, the optimizer type can be set to stochastic gradient descent (SGD), adaptive moment estimation (Adam), root mean square propagation (RMSprop), etc. \cite{DL3}. 

On the other hand, some other hyper-parameters are related to the optimization and training process of DL models; hence, categorized as optimizer hyper-parameters. The learning rate is one of the most important hyper-parameters in DL models \cite{DL4}. It determines the step size at each iteration, which enables the objective function to converge. A large learning rate speeds up the learning process, but the gradient may oscillate around a local minimum value or even cannot converge. On the other hand, a small learning rate converges smoothly, but will largely increase model training time by requiring more training epochs. An appropriate learning rate should enable the objective function to converge to a global minimum in a reasonable amount of time. Another common hyper-parameter is the drop-out rate. Drop-out is a standard regularization method for DL models proposed to reduce over-fitting. In drop-out, a proportion of neurons are randomly removed, and the percentage of neurons to be removed should be tuned. 

Mini-batch size and the number of epochs are the other two DL hyper-parameters that represent the number of processed samples before updating the model, and the number of complete passes through the entire training set, respectively \cite{DL5}. Mini-batch size is affected by the resource requirements of the training process and the number of iterations. The number of epochs depends on the size of the training set and should be tuned by slowly increasing its value until validation accuracy starts to decrease, which indicates over-fitting. On the other hand, DL models often converge within a few epochs, and the following epochs may lead to unnecessary additional execution time and over-fitting, which can be avoided by the early stopping method. Early stopping is a form of regularization whereby model training stops in advance when validation accuracy does not increase after a certain number of consecutive epochs. The number of waiting epochs, called early stop patience, can also be tuned to reduce model training time. 

Apart from traditional DL models, transfer learning (TL) is a technology that obtains a pre-trained model on the data in a related domain and transfers it to other target tasks \cite{TL}. To transfer a DL model from one problem to another problem, a certain number of top layers are frozen, and only the remaining layers are retrained to fit the new problem. Therefore, the number of frozen layers is a vital hyper-parameter to tune if TL is used. 

\subsection{Unsupervised Learning Algorithms}
Unsupervised learning algorithms are a set of ML algorithms used to identify unknown patterns in unlabeled datasets. Clustering and dimensionality-reduction algorithms are the two main types of unsupervised learning methods. Clustering methods include k-means, DBSCAN, EM, hierarchical clustering, etc.; while PCA and LDA are two commonly-used dimensionality reduction algorithms \cite{sklearnbook}.
\subsubsection{Clustering Algorithms}
In most clustering algorithms — including k-means, EM, and hierarchical clustering — the number of clusters is the most important hyper-parameter to tune \cite{ncluster}.

The k-means algorithm \cite{kmeans2} uses $k$ prototypes, indicating the centroids of clusters, to cluster data. In k-means algorithms, the number of clusters, 'n\_clusters', must be specified, and is determined by minimizing the sum of squared errors \cite{kmeans2}:
\begin{equation}
\sum_{i=0}^{n_{k}} \min _{u_{j} \in C_{k}}\left(\mathbf{x}_{i}-u_{j}\right)^{2},
\end{equation}
where $\left(\mathbf{x}_{1}, \cdots, \mathbf{x}_{n}\right)$ is the data matrix; $u_{j}$, also called the centroid of the cluster $C_{k}$, is the mean of the samples in the cluster; and $n_{k}$ is the number of sample points in the cluster $C_{k}$. 

To tune k-means, 'n\_clusters' is the most crucial hyper-parameter. Besides this, the method for centroid initialization, 'init', could be set to 'k-means++', 'random' or a human-defined array, which slightly affects model performance. In addition, 'n\_init', denoting the number of times that the k-means algorithm will be executed with different centroid seeds, and the 'max\_iter', the maximum number of iterations in a single execution of k-means, also have slight impacts on model performance \cite{sklearn}.

The expectation-maximization (EM) algorithm \cite{EM} is an iterative algorithm used to detect the maximum likelihood estimation of parameters. Gaussian Mixture model is a clustering method that uses a mixture of Gaussian distributions to model data by implementing the EM method. Similar to k-means, its major hyper-parameter to be tuned is 'n\_components', indicating the number of clusters or Gaussian distributions. Additionally, different methods can be chosen to constrain the covariance of the estimated classes in Gaussian mixture models, including 'full covariance', 'tied', 'diagonal' or 'spherical' \cite{GMM}. Other hyper-parameters could also be tuned, including 'max\_iter' and 'tol',  representing the number of EM iterations to perform and the convergence threshold, respectively \cite{sklearn}.

Hierarchical clustering \cite{HC} methods build clusters by continuously merging or splitting the built-in clusters. The hierarchy of clusters is represented by a tree-structure; its root indicates the unique cluster gathering all samples, and its leaves represent the clusters with only one sample \cite{HC}. In sklearn, the function 'AgglomerativeClustering' is a common type of hierarchical clustering. In agglomerative clustering, the linkage criteria, 'linkage', determines the distance between sets of observations and can be set to 'ward', 'complete', 'average', or 'single', indicating whether to minimize the variance of the all clusters, or use the maximum, average, or minimum distance between every two clusters, respectively. Like other clustering methods, its main hyper-parameter is the number of clusters, 'n\_clusters'. However, 'n\_clusters' cannot be set if we choose to set the 'distance\_threshold', the linkage distance threshold for merging clusters, since if so, 'n\_clusters' will be determined automatically.

DBSCAN \cite{DBSCAN1} is a density-based clustering method that determines the clusters by dividing data into clusters with sufficiently high density. Unlike other clustering models, the number of clusters does not need to be configured before training. Instead, DBSCAN has two significant conditional hyper-parameters — the scan radius represented by 'eps', and the minimum number of considered neighbor points represented by 'min\_samples' — which define the cluster density together \cite{DBSCAN2}. DBSCAN works by starting with an unvisited point and detecting all its neighbor points within a pre-defined distance 'eps'. If the number of neighbor points reaches the value of 'min\_samples', this unvisited point and all its neighbors are defined as a cluster. The procedures are executed recursively until all data points have been visited. A higher 'min\_samples' or a lower 'eps' indicates a higher density to form a cluster. 

\subsubsection{Dimensionality Reduction Algorithms}
The increasing amount of collected data provides ample information, but also increases problem complexity. In real-world applications, many features are irrelevant or redundant to predict target variables. Dimensionality reduction algorithms often serve as feature engineering methods to extract important features and eliminate insignificant or redundant features. Two common dimensionality-reduction algorithms are principal component analysis (PCA) and linear discriminant analysis (LDA). In PCA and LDA, the number of features to be extracted, represented by 'n\_components' in sklearn, is the main hyper-parameter to be tuned.  

Principal component analysis (PCA) \cite{PCA} is a widely used linear dimensionality reduction method. PCA is based on the concept of mapping the original $n$-dimensional features into $k$-dimension features as the new orthogonal features, also called the principal components. PCA works by calculating the covariance matrix of the data matrix to obtain the eigenvectors of the covariance matrix. The matrix comprises the eigenvectors of $k$ features with the largest eigenvalues (\textit{i.e.}, the largest variance). Consequently, the data matrix can be transformed into a new space with reduced dimensionality. Singular value decomposition (SVD) \cite{SVD} is a popular method used to obtain the eigenvalues and eigenvectors of the covariance matrix of PCA. Therefore, in addition to 'n\_components', the SVD solver type is another hyper-parameter of PCA to be tuned, which can be assigned to 'auto', 'full', 'arpack' or 'randomized' \cite{sklearn}.

Linear discriminant analysis (LDA) \cite{LDA1} is another common dimensionality reduction method that projects the features onto the most discriminative directions. Unlike PCA, which obtains the direction with the largest variance as the principal component, LDA optimizes the feature subspace of classification. The objective of LDA is to minimize the variance inside each class and maximize the variance between different classes after projection. Thus, the projection points in each class should be as close as possible, and the distance between the center points of different classes should be as large as possible. Similar to PCA, the number of features to be extracted, 'n\_components', should be tuned in LDA models. Additionally, the solver type of LDA can also be set to 'svd' for SVD, 'lsqr' for least-squares solution, or 'eigen' for eigenvalue decomposition \cite{LDA2}. LDA also has a conditional hyper-parameter, the shrinkage parameter, 'shrinkage', which can be set to a float value along with 'lsqr' and 'eigen' solvers.

\section{Hyper-parameter Optimization Techniques}
\subsection{Model-free Algorithms}
\subsubsection{Babysitting}
Babysitting, also called 'Trial and Error' or grad student descent (GSD), is a basic hyper-parameter tuning method \cite{ADL}. This method is implemented by 100\% manual tuning and widely used by students and researchers. The workflow is simple: after building a ML model, a student tests many possible hyper-parameter values based on experience, guessing, or the analysis of previously-evaluated results; the process is repeated until this student runs out of time (often reaching a deadline) or is satisfied with the results. As such, this approach requires a sufficient amount of prior knowledge and experience to identify optimal hyper-parameter values with limited time. 

Manual tuning is infeasible for many problems due to several factors, like a large number of hyper-parameters, complex models, time-consuming model evaluations, and non-linear hyper-parameter interactions \cite{BBHPO}. These factors inspired increased research into techniques for the automatic optimization of hyper-parameters \cite{BB2}.

\subsubsection{Grid Search}
Grid search (GS) is one of the most commonly-used methods to explore hyper-parameter configuration space \cite{grid1}. GS can be considered an exhaustive search or a brute-force method that evaluates all the hyper-parameter combinations given to the grid of configurations \cite{grid2}. GS works by evaluating the Cartesian product of a user-specified finite set of values \cite{AMLB}. 

GS cannot exploit the well-performing regions further by itself. Therefore, to identify the global optimums, the following procedure needs to be performed manually \cite{SOAML}:
\begin{enumerate}
\item Start with a large search space and step size.
\item Narrow the search space and step size based on the previous results of well-performing hyper-parameter configurations.
\item Repeat step 2 multiple times until an optimum is reached.
\end{enumerate}

GS can be easily implemented and parallelized. However, the main drawback of GS is its inefficiency for high-dimensionality hyper-parameter configuration space, since the number of evaluations increases exponentially as the number of hyper-parameters grows. This exponential growth is referred to as the curse of dimensionality \cite{Optunity}. For GS, assuming that there are $k$ parameters, and each of them has $n$ distinct values, its computational complexity increases exponentially at a rate of $O(n^k)$ \cite{PSODL}. Thus, only when the hyper-parameter configuration space is small can GS be an effective HPO method. 

\subsubsection{Random Search}
To overcome certain limitations of GS, random search (RS) was proposed in \cite{RS}. RS is similar to GS; but, instead of testing all values in the search space, RS randomly selects a pre-defined number of samples between the upper and lower bounds as candidate hyper-parameter values, and then trains these candidates until the defined budget is exhausted. The theoretical basis of RS is that if the configuration space is large enough, then the global optimums, or at least their approximations, can be detected. With a limited budget, RS is able to explore a larger search space than GS \cite{RS}. 

The main advantage of RS is that it is easily parallelized and resource-allocated since each evaluation is independent. Unlike GS, RS samples a fixed number of parameter combinations from the specified distribution, which improves system efficiency by reducing the probability of wasting much time on a small poor-performing region. Since the number of total evaluations in RS is set to a fixed value $n$ before the optimization process starts, the computational complexity of RS is $O(n)$ \cite{RStime}. In addition, RS can detect the global optimum or the near-global optimum when given enough budgets \cite{AMLB}.

Although RS is more efficient than GS for large search spaces, there are still a large number of unnecessary function evaluations since it does not exploit the previously well-performing regions \cite{SOAML}. 

To conclude, the main limitation of both RS and GS is that every evaluation in their iterations is independent of previous evaluations; thus, they waste massive time evaluating poorly-performing areas of the search space. This issue can be solved by other optimization methods, like Bayesian optimization that uses previous evaluation records to determine the next evaluation \cite{BOHP}.
\subsection{Gradient-based Optimization}
Gradient descent \cite{GBO} is a traditional optimization technique that calculates the gradient of variables to identify the promising direction and moves towards the optimum. After randomly selecting a data point, the technique moves towards the opposite direction of the largest gradient to locate the next data point. Therefore, a local optimum can be reached after convergence. The local optimum is also the global optimum for convex functions.  Gradient-based algorithms have a time complexity of $O(n^k)$ for optimizing $k$ hyper-parameters \cite{GBOtime}. 

For specific machine learning algorithms, the gradient of certain hyper-parameters can be calculated, and then the gradient descent can be used to optimize these hyper-parameters. Although gradient-based algorithms have a faster convergence speed to reach local optimum than the previously-presented methods in Section 4.1, they have several limitations. Firstly, they can only be used to optimize continuous hyper-parameters because other types of hyper-parameters, like categorical hyper-parameters, do not have gradient directions. Secondly, they are only efficient for convex functions because the local instead of a global optimum may be reached for non-convex functions \cite{SOAML}. Therefore, the gradient-based algorithms can only be used in some cases where it is possible to obtain the gradient of hyper-parameters; \textit{e.g.}, optimizing the learning rate in neural networks (NN) \cite{GBad}. Still, it is not guaranteed for ML algorithms to identify global optimums using gradient-based optimization techniques.

\subsection{Bayesian Optimization}
Bayesian optimization (BO) \cite{BO1} is an iterative algorithm that is popularly used for HPO problems. Unlike GS and RS, BO determines the future evaluation points based on the previously-obtained results. To determine the next hyper-parameter configuration, BO   uses two key components: a surrogate model and an acquisition function \cite{RF}. The surrogate model aims to fit all the currently-observed points into the objective function. After obtaining the predictive distribution of the probabilistic surrogate model, the acquisition function determines the usage of different points by balancing the trade-off between exploration and exploitation. Exploration is to sample the instances in the areas that have not been sampled, while exploitation is to sample in the currently promising regions where the global optimum is most likely to occur, based on the posterior distribution. BO models balance the exploration and the exploitation processes to detect the current most likely optimal regions and avoid missing better configurations in the unexplored areas \cite{BO2}. 

The basic procedures of BO are as follows \cite{BO1}:
\begin{enumerate}
\item Build a probabilistic surrogate model of the objective function.
\item Detect the optimal hyper-parameter values on the surrogate model.
\item Apply these hyper-parameter values to the real objective function to evaluate them.
\item Update the surrogate model with new results.
\item Repeat steps 2 - 4 until the maximum number of iterations is reached. 
\end{enumerate}

Thus, BO works by updating the surrogate model after each evaluation on the objective function. BO is more efficient than GS and RS since it can detect the optimal hyper-parameter combinations by analyzing the previously-tested values, and running a surrogate model is often much cheaper than running the entire objective function. 

However, since Bayesian optimization models are executed based on the previously-tested values, they belong to sequential methods that are difficult to parallelize; but they can usually detect near-optimal hyper-parameter combinations within a few iterations \cite{EHPO}.

Common surrogate models for BO include Gaussian process (GP) \cite{GP}, random forest (RF) \cite{SMAC}, and the tree Parzen estimator (TPE) \cite{AHPO}. Therefore, there are three main types of BO algorithms based on their surrogate models: BO-GP, BO-RF, BO-TPE. An alternative name for BO-RF is sequential model-based algorithm configuration (SMAC) \cite{SMAC}. 
\subsubsection{BO-GP}
Gaussian process (GP) is a standard surrogate model for objective function modeling  in BO \cite{BO1}. Assuming that the function $f$ with a mean $\mu$ and a covariance $\sigma^2$ is a realization of a GP, the predictions follow a normal distribution \cite{BOs}: 
\begin{equation}
p(y | x, D)=N\left(y | \hat{\mu}, \hat{\sigma}^{2}\right),
\end{equation}
where $D$ is the configuration space of hyper-parameters, and $y=f(x)$ is the evaluation result of each hyper-parameter value $x$. After obtaining a set of predictions, the points to be evaluated next are then selected from the confidence intervals generated by the BO-GP model. Each newly-tested data point is added to the sample records, and the BO-GP model is re-built with the new information. This procedure is repeated until termination.

 Applying a BO-GP to a size $n$ dataset has a time complexity of $O(n^3)$ and space complexity of $O(n^2)$ \cite{BOGPtime}. One main limitation of BO-GP is that the cubic complexity to the number of instances limits the capacity for parallelization \cite{AMLSC}. Additionally, it is mainly used to optimize continuous variables.
\subsubsection{SMAC}
Random forest (RF) is another popular surrogate function for BO to model the objective function using an ensemble of regression trees. BO using RF as the surrogate model is also called SMAC \cite{SMAC}. 

Assuming that there is a Gaussian model $N\left(y | \hat{\mu}, \hat{\sigma}^{2}\right)$, and $\hat{\mu}$ and $\hat{\sigma}^2$ are the mean and variance of the regression function $r(x)$, respectively, then \cite{SMAC}:
\begin{equation}
\hat{\mu}=\frac{1}{|B|} \sum_{r \in B} r(x),
\end{equation}
\begin{equation}
\hat{\sigma}^{2}=\frac{1}{|B|-1} \sum_{r \in \bar{B}}(r(x)-\hat{\mu})^{2},
\end{equation}
where $B$ is a set of regression trees in the forest. The major procedures of SMAC are as follows \cite{AMLSC}:
\begin{enumerate}
\item RF starts with building $B$ regression trees, each constructed by sampling $n$ instances from the training set with replacement. 
\item A split node is selected from $d$ hyper-parameters for each tree.
\item To maintain a low computational cost, both the minimum number of instances considered for further split and the number of trees to grow are set to a certain value.
\item Finally, the mean and variance for each new configuration are estimated by RF. 
\end{enumerate}

Compared with BO-GP, the main advantage of SMAC is its support for all types of variables, including continuous, discrete, categorical, and conditional hyper-parameters \cite{BOs}. The time complexities of using SMAC to fit and predict variances are $O(nlogn)$ and $O(logn)$, respectively, which are much lower than the complexities of BO-GP \cite{AMLSC}.

\subsubsection{BO-TPE}
Tree-structured Parzen estimator (TPE) \cite{AHPO} is another common surrogate model for BO. Instead of defining a predictive distribution used in BO-GP, BO-TPE creates two density functions, $l(x)$ and $g(x)$, to act as the generative models for all domain variables \cite{AMLSC}. To apply TPE, the observation results are divided into good results and poor results by a pre-defined percentile $y^{*}$, and the two sets of results are modeled by simple Parzen windows \cite{AHPO}:
\begin{equation}
p(x | y, D)=\left\{\begin{array}{ll}{l(x),} & {{ if \quad} y<y^{*}} \\ {g(x),} & { { if \quad} y>y^{*}}\end{array}\right..
\end{equation}

After that, the expected improvement in the acquisition function is reflected by the ratio between the two density functions, which is used to determine the new configurations for evaluation. The Parzen estimators are organized in a tree structure, so the specified conditional dependencies are retained. Therefore, TPE naturally supports specified conditional hyper-parameters \cite{BOs}. The time complexity of BO-TPE is $O(nlogn)$, which is lower than the complexity of BO-GP \cite{AMLSC}.

BO methods are effective for many HPO problems, even if the objective function $f$ is stochastic, non-convex, or non-continuous. However, the main drawback of BO models is that, if they fail to achieve the balance between exploration and exploitation, they might only reach a local instead of a global optimum. RS does not have this limitation since it does not focus on any specific area. Additionally, it is difficult to parallelize BO models since their intermediate results are dependent on each other \cite{EHPO}.
\subsection{Multi-fidelity Optimization Algorithms}
One major issue with HPO is the long execution time, which increases with a larger hyper-parameter configuration space and larger datasets. The execution time may be several hours, several days, or even more \cite{HPS}. Multi-fidelity optimization techniques are common approaches to solve the constraint of limited time and resources. To save time, people can use a subset of the original dataset or a subset of the features \cite{subset}. Multi-fidelity involves low-fidelity and high-fidelity evaluations and combines them for practical applications \cite{multifidelity}. In low-fidelity evaluations, a relatively small subset is evaluated at a low cost but with poor generalization performance. In high-fidelity evaluations, a relatively large subset is evaluated with better generalization performance but at a higher cost than low-fidelity evaluations. In multi-fidelity optimization algorithms, poorly-performing configurations are discarded after each round of hyper-parameter evaluation on generated subsets, and only well-performing hyper-parameter configurations will be evaluated on the entire training set. 

Bandit-based algorithms categorized to multi-fidelity optimization algorithms have shown success dealing with deep learning optimization problems \cite{AMLSC}. Two common bandit-based techniques are successive halving \cite{SH} and Hyperband \cite{Hyperband}. 
\subsubsection{Successive Halving}
Theoretically speaking, exhaustive methods are able to identify the optimal hyper-parameter combination by evaluating all the given combinations. However, many factors, including limited time and resources, should be considered in practical applications. These factors are called budgets ($B$). To overcome the limitations of GS and RS and to improve efficiency, successive halving algorithms were proposed in \cite{SH}. 

The main process of using successive halving algorithms for HPO is as follows. Firstly, it is presumed that there are $n$ sets of hyper-parameter combinations, and that they are evaluated with uniformly-allocated budgets ($b=B/n$). Then, according to the evaluation results for each iteration, half of the poorly-performing hyper-parameter configurations are eliminated, and the better-performing half is passed to the next iteration with double budgets ($b_{i+1}=2 * b_{i}$). The above process is repeated until the final optimal hyper-parameter combination is detected. 

Successive halving is more efficient than RS, but is affected by the trade-off between the number of hyper-parameter configurations and the budgets allocated to each configuration \cite{AMLB}. Thus, the main concern of successive halving is how to allocate the budget and how to determine whether to test fewer configurations with a higher budget for each or to test more configurations with a lower budget for each \cite{SOAML}.
\subsubsection{Hyperband}
Hyperband \cite{Hyperband} is then proposed to solve the dilemma of successive halving algorithms by dynamically choosing a reasonable number of configurations. It aims to achieve a trade-off between the number of hyper-parameter configurations ($n$) and their allocated budgets by dividing the total budgets ($B$) into $n$ pieces and allocating these pieces to each configuration ($b=B/n$). Successive halving serves as a subroutine on each set of random configurations to eliminate the poorly-performing hyper-parameter configurations and improve efficiency. The main steps of Hyperband algorithms are shown in Algorithm 1 \cite{SOAML}.

\begin{algorithm}
\caption{Hyperband}
\begin{algorithmic}[1]
\renewcommand{\algorithmicrequire}{\textbf{Input:}}
\REQUIRE $b_{\max}$, $b_{\min}$
\STATE ${s_{\max}}=\log \left(\frac{b_{\max}}{b_{\min }}\right)$
\FOR {$s \in\left\{b_{\mathrm{max}}, b_{\mathrm{min}}-1, \ldots, 0\right\}$}
\STATE $n=DetermineBudget(s)$
\STATE $\gamma=SampleConfigurations(n)$
\STATE $SuccessiveHalving(\gamma)$
\ENDFOR
\RETURN \textit{The best configuration so far.}
\end{algorithmic} 
\end{algorithm}

Firstly, the budget constraints $b_{min}$ and $b_{max}$ are determined by the total number of data points, the minimum number of instances required to train a sensible model, and the available budgets. After that, the number of configurations $n$ and the budget size allocated to each configuration are calculated based on $b_{min}$ and $b_{max}$ in steps 2-3 of Algorithm 1. The configurations are sampled based on $n$ and $b$, and then passed to the successive halving model demonstrated in steps 4-5. The successive halving algorithm discards the identified poorly-performing configurations and passes the well-performing configurations on to the next iteration. This process is repeated until the final optimal hyper-parameter configuration is identified.  By involving the successive halving searching method, Hyperband has a computational complexity of $O(nlogn)$ \cite{Hyperband}.
\subsubsection{BOHB}
Bayesian Optimization HyperBand (BOHB) \cite{BOHB} is a state-of-the-art HPO technique that combines Bayesian optimization and Hyperband to incorporate the advantages of both while avoiding their drawbacks. The original Hyperband uses a random search to search the hyper-parameter configuration space, which has a low efficiency. BOHB replaces the RS method by BO to achieve both high performance as well as low execution time by effectively using parallel resources to optimize all types of hyper-parameters. In BOHB, TPE is the standard surrogate model for BO, but it uses multidimensional kernel density estimators. Therefore, the complexity of BOHB is also $O(nlogn)$ \cite{BOHB}.

It has been shown that BOHB outperforms many other optimization techniques when tuning SVM and DL models \cite{BOHB}. The only limitation of BOHB is that it requires the evaluations on subsets with small budgets to be representative of evaluations on the entire training set; otherwise, BOHB may have a slower convergence speed than standard BO models. 
\subsection{Metaheuristic Algorithms}
Metaheuristic algorithms \cite{Metaheuristic} are a set of algorithms mainly inspired by biological theories and widely used for optimization problems. Unlike many traditional optimization methods, metaheuristics have the capacity to solve non-convex, non-continuous, and non-smooth optimization problems. 

Population-based optimization algorithms (POAs) are a major type of metaheuristic algorithm, including genetic algorithms (GAs), evolutionary algorithms, evolutionary strategies, and particle swarm optimization (PSO). 
POAs start by creating and updating a population as each generation; each individual in every generation is then evaluated until the global optimum is identified \cite{BOHP}. The main differences between different POAs are the methods used to generate and select populations \cite{HumanAML}. POAs can be easily parallelized since a population of $N$ individuals can be evaluated on at most $N$ threads or machines in parallel \cite{AMLB}. Genetic algorithms and particle swarm optimization are the two main POAs that are popularly-used for HPO problems.
\subsubsection{Genetic Algorithm}
Genetic algorithm (GA) \cite{GA2} is one of the common metaheuristic algorithms based on the evolutionary theory that individuals with the best survival capability and adaptability to the environment are more likely to survive and pass on their capabilities to future generations. The next generation will also inherit their parents' characteristics and may involve better and worse individuals. Better individuals will be more likely to survive and have more capable offspring, while the worse individuals will gradually disappear. After several generations, the individual with the best adaptability will be identified as the global optimum \cite{GA3}.

To apply GA to HPO problems, each chromosome or individual represents a hyper-parameter, and its decimal value is the actual input value of the hyper-parameter in each evaluation. Every chromosome has several genes, which are binary digits; and then crossover and mutation operations are performed on the genes of this chromosome. The population involves all possible values within the initialized chromosome/parameter ranges, while the fitness function characterizes the evaluation metrics of the parameters \cite{GA3}.

Since the randomly-initialized parameter values often do not include the optimal parameter values, several operations, including selection, crossover, and mutation operations, must be performed on the well-performing chromosomes to identify the optimums \cite{GA2}. Chromosome selection is implemented by selecting those chromosomes with good fitness function values. To keep the population size unchanged, the chromosomes with good fitness function values are passed to the next generation with higher probability, where they generate new chromosomes with the parents' best characteristics. Chromosome selection ensures that good characteristics of each generation can be passed to later generations. Crossover is used to generate new chromosomes by exchanging a proportion of genes in different chromosomes. Mutation operations are also used to generate new chromosomes by randomly altering one or more genes of a chromosome. Crossover and mutation operations enable later generations to have different characteristics and reduce the chance of missing good characteristics \cite{AMLSC}. 

The main procedures of GA are as follows \cite{Metaheuristic}: 
\begin{enumerate}
\item Randomly initialize the population, chromosomes, and genes, which represent the entire search space, hyper-parameters, and hyper-parameter values, respectively.
\item Evaluate the performance of each individual in the current generation by calculating the fitness function, which indicates the objective function of a ML model. 
\item Perform selection, crossover, and mutation operations on the chromosomes to produce a new generation involving the next hyper-parameter configurations to be evaluated.
\item Repeat steps 2 \& 3 until the termination condition is met.
\item Terminate and output the optimal hyper-parameter configuration.
\end{enumerate}

Among the above steps, the population initialization step is an important step of GA and PSO since it provides an initial guess of the optimal values. Although the initialized values will be iteratively improved in the optimization process, a suitable population initialization method can significantly improve the convergence speed and performance of POAs. A good initial population of hyper-parameters should involve individuals that are close to global optimums by covering the promising regions and should not be localized to an unpromising region of the search space \cite{goodini}. 

To generate hyper-parameter configuration candidates for the initial population, random initialization that simply creates the initial population with random values in the given search space is often used in GA \cite{ini1}. Thus, GA is easily implemented and does not necessitate good initializations, because its selection, crossover, and mutation operations lower the possibility of missing the global optimum.

Hence, it is useful when the data analyst does not have much experience determining a potential appropriate initial search space for the hyper-parameters. The main limitation of GA is that the algorithm itself introduces additional hyper-parameters to be configured, including the fitness function type, population size, crossover rate, and mutation rate. Moreover, GA is a sequential execution algorithm, making it difficult to parallelize. The time complexity of GA is $O(n^2)$ \cite{GAtime}. As a result, sometimes, GA may be inefficient due to its low convergence speed. 
\subsubsection{Particle Swarm Optimization}
Particle swarm optimization (PSO) \cite{PSO1} is another set of evolutionary algorithms that are commonly used for optimization problems. PSO algorithms are inspired by biological populations that exhibit both individual and social behaviors \cite{HumanAML}. PSO works by enabling a group of particles (swarm) to traverse the search space in a semi-random manner \cite{BBHPO}. PSO algorithms identify the optimal solution through cooperation and information sharing among individual particles in a group. 

In PSO, there are a group of $n$ particles in a swarm $\mathbf{S}$ \cite{SOAML}: 
\begin{equation}
\mathbf{S}=\left(S_{1}, S_{2}, \cdots, S_{n}\right),
\end{equation}
and each particle $S_{i}$ is represented by a vector: 
\begin{equation}
S_{i}=<\overrightarrow{x_{i}}, \overrightarrow{v_{i}}, \overrightarrow{p_{i}}>,
\end{equation}
where $\overrightarrow{x_{i}}$ is the current position, $\overrightarrow{v_{i}}$ is the current velocity, and $\overrightarrow{p_{i}}$ is the known best position of the particle so far.

After initializing the position and velocity of each particle,very particle evaluates the current position and records the position with its performance score. In the next iteration, the velocity $\overrightarrow{v_{i}}$ of each particle is changed based on the previous position $\overrightarrow{p_{i}}$ and the current global optimal position $\overrightarrow{p}$:
\begin{equation}
\overrightarrow{v_{i}}:=\overrightarrow{v_{i}}+U\left(0, \varphi_{1}\right)(\overrightarrow{p_{i}}-\overrightarrow{x_{i}})+U\left(0, \varphi_{2}\right)(\overrightarrow{p}-\overrightarrow{x_{i}}),
\end{equation}
where $U(0, \varphi)$ is the continuous uniform distributions based on the acceleration constants $\varphi_{1}$ and $\varphi_{2}$.

After that, the particles move based on their new velocity vectors:
\begin{equation}
\overrightarrow{x_{i}}:=\overrightarrow{x_{i}}+\overrightarrow{v_{i}}.
\end{equation}

The above procedures are repeated until convergence or termination constraints are reached.

Compared with GA, it is easier to implement PSO, since PSO does not have certain additional operations like crossover and mutation. In GA, all chromosomes share information with each other, so the entire population moves uniformly toward the optimal region; while in PSO, only information on the individual best particle and the global best particle is transmitted to others, which is a one-way flow of information sharing, and the entire search process follows the direction of the current optimal solution \cite{SOAML}. The computational complexity of PSO algorithm is $O(nlogn)$ \cite{PSOtime}. In most cases, the convergence speed of PSO is faster than of GA. In addition, particles in PSO operate independently and only need to share information with each other after each iteration, so this process is easily parallelized to improve model efficiency \cite{BBHPO}. 

The main limitation of PSO is that it requires proper population initialization; otherwise, it might only reach a local instead of a global optimum, especially for discrete hyper-parameters \cite{PSOdiscrete}.  Proper population initialization requires developers’ prior experience or using population initialization techniques. Many population initialization techniques have been proposed to improve the performance of evolutionary algorithms, like the opposition-based optimization algorithm \cite{ini1} and the space transformation search method \cite{ini2}. Involving additional population initialization techniques will require more execution time and resources.

\section{Applying Optimization Techniques to Machine Learning Algorithms}
\subsection{Optimization Techniques Analysis}
Grid search (GS) is a simple method, its major limitation being that it is time-consuming and impacted by the curse of dimensionality \cite{Optunity}. Thus, it is unsuitable for a large number of hyper-parameters. Moreover, GS is often not able to detect the global optimum of continuous parameters, since it requires a pre-defined, finite set of hyper-parameter values. It is also unrealistic for GS to be used to identify integer and continuous hyper-parameter optimums with limited time and resources. Therefore, compared with other techniques, GS is only efficient for a small number of categorical hyper-parameters.
 
Random search is more efficient than GS and supports all types of hyper-parameters. In practical applications, using RS to evaluate the randomly-selected hyper-parameter values helps analysts to explore a large search space. However, since RS does not consider previously-tested results, it may involve many unnecessary evaluations, which decrease its efficiency \cite{RS}. 

Hyperband can be considered an improved version of RS, and they both support parallel executions. Hyperband balances model performance and resource usage, so it is more efficient than RS, especially with limited time and resources \cite{surrogates}. However, GS, RS, and Hyperband all have a major constraint in that they treat each hyper-parameter independently and do not consider hyper-parameter correlations \cite{HBBO}. Thus, they will be inefficient for ML algorithms with conditional hyper-parameters, like SVM, DBSCAN, and logistic regression.

Gradient-based algorithms are not a prevalent choice for hyper-parameter optimization, since they only support continuous hyper-parameters and can only detect a local instead of a global optimum for non-convex HPO problems \cite{SOAML}. Therefore, gradient-based algorithms can only be used to optimize certain hyper-parameters, like the learning rate in DL models.

Bayesian optimization models are divided into three different models — BO-GP, SMAC, and BO-TPE — based on their surrogate models. BO algorithms determine the next hyper-parameter value based on the previously-evaluated results to reduce unnecessary evaluations and improve efficiency. BO-GP mainly supports continuous and discrete hyper-parameters (by rounding them), but does not support conditional hyper-parameters \cite{BOHP}; while SMAC and BO-TPE are both able to handle categorical, discrete, continuous, and conditional hyper-parameters. SMAC performs better when there are many categorical and conditional parameters, or cross-validation is used, while BO-GP performs better for only a few continuous parameters \cite{surrogates}. BO-TPE preserves the specified conditional relationships, so one advantage of BO-TPE over BO-GP is its innate support for specified conditional hyper-parameters \cite{BOHP}.

Metaheuristic algorithms, including GA and PSO, are more complicated than many other HPO algorithms, but often perform well for complex optimization problems. They support all types of hyper-parameters and are particularly efficient for large configuration spaces, since they can obtain the near-optimal solutions even within very few iterations. However, GA and PSO have their own advantages and disadvantages in practical use. PSO is able to support large-scale parallelization, and is particularly suitable for continuous and conditional HPO problems \cite{PSODL}; on the other hand, GA is executed sequentially, making it difficult to be parallelized. Therefore, PSO often executes faster than GA, especially for large configuration spaces and large datasets. However, an appropriate population initialization is crucial for PSO; otherwise, it may converge slowly or only identify a local instead of a global optimum. Yet, the impact of proper population initialization is not as significant for GA as for PSO \cite{PSO3}. Another limitation of GA is that it introduces additional hyper-parameters, like its crossover and mutation rates \cite{GA2}.

The strengths and limitations of the hyper-parameter optimization algorithms involved in this paper are summarized in Table \ref{t1}.

\begin{table*}[!ht]
\caption{The comparison of common HPO algorithms ($n$ is the number of hyper-parameter values and $k$ is the number of hyper-parameters)}
\setlength\extrarowheight{1pt}
\centering
\scriptsize
\begin{tabular}{p{1.35cm}|p{4.25cm}|p{4.4cm}|p{1.25cm}}
\Xhline{1.2pt}

\hline
\textbf{HPO Method} & \textbf{Strengths }                                                                                            & \textbf{Limitations }                                                                                                                       & \textbf{Time Complexity}  \\ \Xhline{1.2pt}
\hline
GS                           & ·         Simple.                                                                                              & \begin{tabular}[c]{@{}p{4.4cm}}· Time-consuming,\\ · Only efficient with categorical HPs. \end{tabular}                                         & $O(n^k)$                        \\ 
\hline
RS                           & \begin{tabular}[c]{@{}p{4.25cm}}· More efficient than GS.\\ · Enable parallelization.\end{tabular}                  & \begin{tabular}[c]{@{}p{4.4cm}}· Not consider previous results.\\ · Not efficient with conditional HPs. \end{tabular}                           & $O(n)$                          \\ 
\hline
Gradient-based models        & \multicolumn{1}{l|}{\multirow{2}{*}{ \begin{tabular}[c]{@{}p{4.25cm}}·         Fast convergence speed for continuous HPs.     \end{tabular} }}                                                 & \multicolumn{1}{l|}{\multirow{2}{*}{\begin{tabular}[c]{@{}p{4.4cm}}· Only support continuous HPs.\\ · May only detect local optimums. \end{tabular}    }}                             & \multicolumn{1}{l}{\multirow{2}{*}{$O(n^{k})$ }}                       \\ 
\hline
BO-GP                      &  \begin{tabular}[c]{@{}p{4.25cm}}·         Fast convergence speed for continuous HPs.     \end{tabular}                                                         & \begin{tabular}[c]{@{}p{4.4cm}}· Poor capacity for parallelization.\\· Not efficient with conditional HPs.\end{tabular}                         & $O(n^{3})$                      \\ 
\hline
SMAC                         & ·         Efficient with all types of HPs.                                                                      & ·         Poor capacity for parallelization.                                                                                                & $O(nlogn)$                       \\ 
\hline
BO-TPE                       & \begin{tabular}[c]{@{}p{4.25cm}}· Efficient with all types of HPs.\\ · Keep conditional dependencies. \end{tabular}  & ·         Poor capacity for parallelization.                                                                                                & $O(nlogn)$                       \\ 
\hline
Hyperband                    & ·         Enable parallelization.                                                                              & \begin{tabular}[c]{@{}p{4.4cm}}· Not efficient with conditional HPs.\\ · Require subsets with small budgets to be representative. \end{tabular} & $O(nlogn)$                       \\ 
\hline
\multicolumn{1}{l|}{\multirow{2}{*}{BOHB}}                  & \multirow{2}{*}{\begin{tabular}[c]{@{}p{4.25cm}}·         Efficient with all types of HPs.\\ ·         Enable parallelization.\end{tabular}}       & {· Require subsets with small budgets to be representative.}                                                                          & \multicolumn{1}{l}{\multirow{2}{*}{$O(nlogn)$ }}                      \\ 
\hline
GA                           & \begin{tabular}[c]{@{}p{4.25cm}}· Efficient with all types of HPs.\\ · Not require good initialization.\end{tabular} & ·         Poor capacity for parallelization.                                                                                                & $O(n^{2})$                      \\ 
\hline
PSO                          & \begin{tabular}[c]{@{}p{4.25cm}}· Efficient with all types of HPs.\\ · Enable parallelization.\end{tabular}          & · Require~proper initialization.                                                                                                            & $O(nlogn)$                       \\
\Xhline{1.2pt}
\end{tabular}
\label{t1}%
\end{table*}

\subsection{Apply HPO Algorithms to ML Models}
Since there are many different HPO methods for different use cases, it is crucial to select the appropriate optimization techniques for different ML models. 

Firstly, if we have access to multiple fidelities, which means that it is able to define meaningful budgets: the performance rankings of hyper-parameter configurations evaluated on small budgets should be the same as or similar to the configuration rankings on the full budget (the original dataset); BOHB would be the best choice, since it has the advantages of both BO and Hyperband \cite{AMLB} \cite{BOHB}.

On the other hand, if multiple fidelities are not applicable, which means that using the subsets of the original dataset or the subsets of original features is misleading or too noisy to reflect the performance of the entire dataset, BOHB may perform poorly with higher time complexity than standard BO models, then choosing other HPO algorithms would be more efficient \cite{BOHB}. 

ML algorithms can be classified by the characteristics of their hyper-parameter configurations. Appropriate optimization algorithms can be chosen to optimize the hyper-parameters based on these characteristics.

\subsubsection{One Discrete Hyper-parameter}
Commonly for some ML algorithms, like certain neighbor-based, clustering, and dimensionality reduction algorithms, only one discrete hyper-parameter needs to be tuned. For KNN, the major hyper-parameter is $k$, the number of considered neighbors. The most essential hyper-parameter of k-means, hierarchical clustering, and EM is the number of clusters. Similarly, for dimensionality reduction algorithms, including PCA and LDA, their basic hyper-parameter is 'n\_components', the number of features to be extracted. 

In these situations, Bayesian optimization is the best choice, and the three surrogates could be tested to find the best one. Hyperband is another good choice, which may have a fast execution speed due to its capacity for parallelization. In some cases, people may want to fine-tune the ML model by considering other less important hyper-parameters, like the distance metric of KNN and the SVD solver type of PCA; so BO-TPE, GA, or PSO could be chosen for these situations.
\subsubsection{One Continuous Hyper-parameter} 
Some linear models, including ridge and lasso algorithms, and some naïve Bayes algorithms, involving multinomial NB, Bernoulli NB, and complement NB, generally only have one vital continuous hyper-parameter to be tuned. In ridge and lasso algorithms, the continuous hyper-parameter is 'alpha', the regularization strength. In the three NB algorithms mentioned above, the critical hyper-parameter is also named 'alpha', but it represents the additive (Laplace/Lidstone) smoothing parameter. In terms of these ML algorithms, BO-GP is the best choice, since it is good at optimizing a small number of continuous hyper-parameters. Gradient-based algorithms can also be used, but might only detect local optimums, so they are less effective than BO-GP. 
\subsubsection{A Few Conditional Hyper-parameters}
It is noticeable that many ML algorithms have conditional hyper-parameters, like SVM, LR, and DBSCAN. LR has three correlated hyper-parameters, 'penalty', '$C$', and the solver type. Similarly, DBSCAN has 'eps' and 'min\_samples' that must be tuned in conjunction. SVM is more complex, since after setting a different kernel type, there is a separate set of conditional hyper-parameters that need to be tuned next, as described in Section 3.1.3. Hence, some HPO methods that cannot effectively optimize conditional hyper-parameters, including GS, RS, BO-GP, and Hyperband, are not suitable for ML models with conditional hyper-parameters. For these ML methods, BO-TPE is the best choice if we have pre-defined relationships among the hyper-parameters. SMAC is also a good choice, since it also performs well for tuning conditional hyper-parameters. GA and PSO can be used, as well.
\subsubsection{A Large Hyper-parameter Configuration Space with Multiple Types of Hyper-parameters}
Tree-based algorithms, including DT, RF, ET, and XGBoost, as well as DL algorithms, like DNN, CNN, RNN, are the most complex types of ML algorithms to bed tuned, since they have many hyper-parameters with various, different types. For these ML models, PSO is the best choice since it enables parallel executions to improve efficiency, particularly for DL models that often require massive training time. Some other techniques, like GA, BO-TPE, and SMAC can also be used, but they may cost more time than PSO, since it is difficult to parallelize these techniques.
\subsubsection{Categorical Hyper-parameters}
This category of hyper-parameters is mainly for ensemble learning algorithms, since their major hyper-parameter is a categorical hyper-parameter. For bagging and AdaBoost, the categorical hyper-parameter is 'base\_estimator', which is set to be a singular ML model. For voting, it is 'estimators', indicating a list of ML singular models to be combined. The voting method has another categorical hyper-parameter, 'voting', which is used to choose whether to use a hard or soft voting method. If we only consider these categorical hyper-parameters, GS would be sufficient to detect their suitable base machine learners. On the other hand, in many cases, other hyper-parameters need to be considered, like 'n\_estimators', 'max\_samples', and 'max\_features' in bagging, as well as 'n\_estimators' and 'learning\_rate' in AdaBoost; consequently, BO algorithms would be a better choice to optimize these continuous or discrete hyper-parameters.  

In conclusion, when tuning a ML model to achieve high model performance and low computational costs, the most suitable HPO algorithm should be selected based on the properties of its hyper-parameters.

\section{Existing HPO Frameworks}
To tackle HPO problems, many open-source libraries exist to apply theory into practice and lower the threshold for ML developers. In this section, we provide a brief introduction to some popular open-source HPO libraries or frameworks mainly for Python programming. The principles behind the involved optimization algorithms are provided in Section 4. 
\subsection{Sklearn}
In sklearn \cite{sklearn}, 'GridSearchCV' can be implemented to detect the optimal hyper-parameters using the GS algorithm. Each hyper-parameter value in the human-defined configuration space is evaluated by the program, with its performance evaluated using cross-validation. When all the instances in the configuration space have been evaluated, the optimal hyper-parameter combination in the defined search space with its performance score will be returned. 

'RandomizedSearchCV' is also provided in sklearn to implement a RS method. It evaluates a pre-defined number of randomly-selected hyper-parameter values in parallel. Cross-validation is conducted to effectively evaluate the performance of each configuration.
\subsection{Spearmint}
Spearmint \cite{BO1} is a library using Bayesian optimization with the Gaussian process as the surrogate model. Spearmint's primary deficiency is that it is not very efficient for categorical and conditional hyper-parameters.
\subsection{BayesOpt}
Bayesian Optimization (BayesOpt) \cite{BayesOpt} is a Python library employed to solve HPO problems using BO. BayesOpt uses a Gaussian process as its surrogate model to calculate the objective function based on past evaluations and utilizes an acquisition function to determine the next values. 
\subsection{Hyperopt}
Hyperopt \cite{Hyperopt} is a HPO framework that involves RS and BO-TPE as the optimization algorithms. Unlike some of the other libraries that only support a single model, Hyperopt is able to use multiple models to model hierarchical hyper-parameters. In addition, Hyperopt is parallelizable since it uses MongoDb as the central database to store the hyper-parameter combinations. Hyperopt-sklearn \cite{hyperopt-sklearn} and hyperas \cite{hyperas} are the two libraries that can apply Hyperopt to scikit-learn and Keras libraries.
\subsection{SMAC}
SMAC \cite{SMAC}\cite{SMAC3} is another library that uses BO with random forest as the surrogate model. It supports categorical, continuous, and discrete variables. 
\subsection{BOHB}
BOHB framework \cite{BOHB} is a combination of Bayesian optimization and Hyperband \cite{surrogates}. It overcomes one limitation of Hyperband, in that it randomly generates the test configurations, by replacing this procedure by BO. TPE is used as the surrogate model to store and model function evaluations. Using BOHB to evaluate the instance can achieve a trade-off between model performance and the current budget. 
\subsection{Optunity}
Optunity \cite{Optunity} is a popular HPO framework that provides several optimization techniques, including GS, RS, PSO, and BO-TPE. In Optunity, categorical hyper-parameters are converted to discrete hyper-parameters by indexing, and discrete hyper-parameters are processed as continuous hyper-parameters by rounding them; as such, it supports all types of hyper-parameters.
\subsection{Skopt}
Skopt (scikit-optimize) \cite{SKOPT} is a HPO library that is built on top of the scikit-learn \cite{sklearn} library. It implements several sequential model-based optimization models, including RS and BO-GP. The methods exhibit good performance with small search space and proper initialization.
\subsection{GpFlowOpt}
GpFlowOpt \cite{GPflowOpt} is a Python library for BO using GP as the surrogate model. It supports running BO-GP on GPU using the Tensorflow library. Therefore, GpFlowOpt is a good choice if BO is used in deep learning models with GPU resources available.
\subsection{Talos}
Talos \cite{talos} is a Python package designed for hyper-parameter optimization with Keras models. Talos can be fully deployed into any Keras models and implemented easily without learning any new syntax. Several optimization techniques, including GS, RS, and probabilistic reduction, can be implemented using Talos.
\subsection{Sherpa}
Sherpa \cite{SHERPA} is a Python package used for HPO problems. It can be used with other ML libraries, including sklearn \cite{sklearn}, Tensorflow\cite{tf}, and Keras \cite{keras}. It supports parallel computations and has several optimization methods, including GS, RS, BO-GP (via GPyOpt), Hyperband, and population-based training (PBT). 
\subsection{Osprey}
Osprey \cite{Osprey} is a Python library designed to optimize hyper-parameters. Several HPO strategies are available in Osprey, including GS, RS, BO-TPE (via Hyperopt), and BO-GP (via GPyOpt). 
\subsection{FAR-HO}
FAR-HO \cite{FAR-HO} is a hyper-parameter optimization package that employs gradient-based algorithms with TensorFlow. FAR-HO contains a few gradient-based optimizers, like reverse hyper-gradient and forward hyper-gradient methods. This library is designed to build access to the gradient-based hyper-parameter optimizers in TensorFlow, allowing deep learning model training and hyper-parameter optimization in GPU or other tensor-optimized computing environments.
\subsection{Hyperband}
Hyperband \cite{Hyperband} is a Python package for tuning hyper-parameters by Hyperband, a bandit-based approach. Similar to 'GridSearchCV' and 'RandomizedSearchCV' in scikit-learn, there is a class named 'HyperbandSearchCV' in Hyperband that can be combined with sklearn and used for HPO problems. In 'HyperbandSearchCV' method, cross-validation is used for evaluation.
\subsection{DEAP}
DEAP \cite{DEAP} is a novel evolutionary computation package for Python that contains several evolutionary algorithms like GA and PSO. It integrates with parallelization mechanisms like multiprocessing, and machine learning packages like sklearn.
\subsection{TPOT}
TPOT \cite{TPOT} is a Python tool for auto-ML that uses genetic programming to optimize ML pipelines. TPOT is built on top of sklearn, so it is easy to implement TPOT on ML models. 'TPOTClassifier' is its principal function, and several additional hyper-parameters of GA must be set to fit specific problems.
\subsection{Nevergrad}
Nevergrad \cite{nevergrad} is an open-source Python library that includes a wide range of optimizers, like fast-GA and PSO. In ML, Nevergrad can be used to tune all types of hyper-parameters, including discrete, continuous, and categorical hyper-parameters, by choosing different optimizers.

\section{Experiments}
To summarize the content of Sections 3 to 6, a comprehensive overview of applying hyper-parameter optimization techniques to ML models is shown in Table \ref{t2o}. It provides a summary of common ML algorithms, their hyper-parameters, suitable optimization methods, and available Python libraries; thus, data analysts and researchers can look up this table and select suitable optimization algorithms as well as libraries for practical use.

\begin{table*}[]
\caption{A comprehensive overview of common ML models, their hyper-parameters, suitable optimization techniques, and available Python libraries}
\setlength\extrarowheight{0.1pt}
\centering
\tiny
\begin{tabular}{c|c|c|c|c}
\Xhline{1.2pt}
 \multicolumn{1}{c|}{\textbf{ML Algorithm}} & \textbf{Main HPs}                                                                                                                                                                                         & \textbf{Optional HPs}                                                                               & \textbf{HPO methods}                                            & \multicolumn{1}{c}{\textbf{Libraries}}                                                  \\ \Xhline{1.2pt}
Linear regression                           & -                                                                                                                                                                                                         & -                                                                                                   & -                                                                           & -                                                                                                   \\ \hline
Ridge \& lasso                                & alpha                                                                                                                                                                                                     & -                                                                                                   & BO-GP                                                                       & Skpot                                                                                               \\ \hline

Logistic regression                         & \begin{tabular}[c]{@{}c@{}}penalty,\\ c,\\ solver\end{tabular}                                                                                                                                            & -                                                                                                   & \begin{tabular}[c]{@{}c@{}}BO-TPE,\\ SMAC\end{tabular}                      & \begin{tabular}[c]{@{}c@{}}Hyperopt,\\ SMAC\end{tabular}                                            \\ \hline
KNN                                           & n\_neighbors                                                                                                                                                                                              & \begin{tabular}[c]{@{}c@{}}weights,\\ p,\\ algorithm\end{tabular}                                   & \begin{tabular}[c]{@{}c@{}}BOs,\\ Hyperband\end{tabular}                    & \begin{tabular}[c]{@{}c@{}}Skpot,\\ Hyperopt,\\ SMAC,\\ Hyperband\end{tabular}                      \\ \hline

SVM                                         & \begin{tabular}[c]{@{}c@{}}C,\\ kernel,\\ epsilon (for SVR)\end{tabular}                                                                                                                                  & \begin{tabular}[c]{@{}c@{}}gamma,\\ coef0,\\ degree\end{tabular}                                    & \begin{tabular}[c]{@{}c@{}}BO-TPE,\\ SMAC,\\ BOHB\end{tabular}              & \begin{tabular}[c]{@{}c@{}}Hyperopt,\\ SMAC,\\ BOHB\end{tabular}                                    \\ \hline
NB                                            & alpha                                                                                                                                                                                                     & -                                                                                                   & BO-GP                                                                       & Skpot                                                                                               \\ \hline

DT                                          & \begin{tabular}[c]{@{}c@{}}criterion,\\ max\_depth,\\ min\_samples\_split,\\ min\_samples\_leaf,\\ max\_features\end{tabular}                                                                             & \begin{tabular}[c]{@{}c@{}}splitter,\\ min\_weight\_fraction\_leaf,\\ max\_leaf\_nodes\end{tabular} & \begin{tabular}[c]{@{}c@{}}GA,\\ PSO,\\ BO-TPE,\\ SMAC,\\ BOHB\end{tabular} & \begin{tabular}[c]{@{}c@{}}TPOT,\\ Optunity,\\ SMAC,\\ BOHB\end{tabular}                            \\ \hline
RF \& ET                                        & \begin{tabular}[c]{@{}c@{}}n\_estimators\\ max\_depth,\\ criterion,\\ min\_samples\_split,\\ min\_samples\_leaf,\\ max\_features\end{tabular}                                                             & \begin{tabular}[c]{@{}c@{}}splitter,\\ min\_weight\_fraction\_leaf,\\ max\_leaf\_nodes\end{tabular} & \begin{tabular}[c]{@{}c@{}}GA,\\ PSO,\\ BO-TPE,\\ SMAC,\\ BOHB\end{tabular} & \begin{tabular}[c]{@{}c@{}}TPOT,\\ Optunity,\\ SMAC,\\ BOHB\end{tabular}                            \\ \hline

XGBoost                                     & \begin{tabular}[c]{@{}c@{}}n\_estimators,\\ max\_depth,\\ learning\_rate,\\ subsample,\\ colsample\_bytree,\end{tabular}                                                                                  & \begin{tabular}[c]{@{}c@{}}min\_child\_weight,\\ gamma,\\ alpha,\\ lambda\end{tabular}              & \begin{tabular}[c]{@{}c@{}}GA,\\ PSO,\\ BO-TPE,\\ SMAC,\\ BOHB\end{tabular} & \begin{tabular}[c]{@{}c@{}}TPOT,\\ Optunity,\\ SMAC,\\ BOHB\end{tabular}                            \\ \hline
Voting                                        & \begin{tabular}[c]{@{}c@{}}estimators,\\ voting\end{tabular}                                                                                                                                              & weights                                                                                             & GS                                                                          & sklearn                                                                                             \\ \hline

Bagging                                     & \begin{tabular}[c]{@{}c@{}}base\_estimator,\\ n\_estimators\end{tabular}                                                                                                                                  & \begin{tabular}[c]{@{}c@{}}max\_samples,\\ max\_features\end{tabular}                               & \begin{tabular}[c]{@{}c@{}}GS,\\ BOs\end{tabular}                           & \begin{tabular}[c]{@{}c@{}}sklearn,\\ Skpot,\\ Hyperopt,\\ SMAC\end{tabular}                        \\ \hline
AdaBoost                                      & \begin{tabular}[c]{@{}c@{}}base\_estimator,\\ n\_estimators,\\ learning\_rate\end{tabular}                                                                                                                & -                                                                                                   & \begin{tabular}[c]{@{}c@{}}BO-TPE,\\ SMAC\end{tabular}                      & \begin{tabular}[c]{@{}c@{}}Hyperopt,\\ SMAC\end{tabular}                                            \\ \hline

Deep learning                               & \begin{tabular}[c]{@{}c@{}}number of hidden layers,\\ ‘units’ per layer,\\ loss,\\ optimizer,\\ Activation,\\ learning\_rate,\\ dropout rate,\\ epochs,\\ batch\_size,\\ early stop patience\end{tabular} & \begin{tabular}[c]{@{}c@{}}number of frozen layers\\ (if transfer learning\\ is used)\end{tabular}    & \begin{tabular}[c]{@{}c@{}}PSO,\\ BOHB\end{tabular}                         & \begin{tabular}[c]{@{}c@{}}Optunity,\\ BOHB\end{tabular}                                            \\ \hline
K-means                                       & n\_clusters                                                                                                                                                                                               & \begin{tabular}[c]{@{}c@{}}init,\\ n\_init,\\ max\_iter\end{tabular}                                & \begin{tabular}[c]{@{}c@{}}BOs,\\ Hyperband\end{tabular}                    & \begin{tabular}[c]{@{}c@{}}Skpot,\\ Hyperopt,\\ SMAC,\\ Hyperband\end{tabular}                      \\ \hline

Hierarchical clustering                     & \begin{tabular}[c]{@{}c@{}}n\_clusters,\\ distance\_threshold\end{tabular}                                                                                                                                & linkage                                                                                             & \begin{tabular}[c]{@{}c@{}}BOs,\\ Hyperband\end{tabular}                    & \begin{tabular}[c]{@{}c@{}}Skpot,\\ Hyperopt,\\ SMAC,\\ Hyperband\end{tabular}                      \\ \hline
DBSCAN                                        & \begin{tabular}[c]{@{}c@{}}eps,\\ min\_samples\end{tabular}                                                                                                                                               & -                                                                                                   & \begin{tabular}[c]{@{}c@{}}BO-TPE,\\ SMAC,\\ BOHB\end{tabular}              & \begin{tabular}[c]{@{}c@{}}Hyperopt,\\ SMAC,\\ BOHB\end{tabular}                                    \\ \hline

Gaussian mixture                       & n\_components                                                                                                                                                                                             & \begin{tabular}[c]{@{}c@{}}covariance\_type,\\ max\_iter,\\ tol\end{tabular}                        & BO-GP                                                                       & Skpot                                                                                               \\ \hline
PCA                                           & n\_components                                                                                                                                                                                             & svd\_solver                                                                                         & \begin{tabular}[c]{@{}c@{}}BOs,\\ Hyperband\end{tabular}                    & \begin{tabular}[c]{@{}c@{}}Skpot,\\ Hyperopt,\\ SMAC,\\ Hyperband\end{tabular}                      \\ \hline

LDA                                           & n\_components                                                                                                                                                                                             & \begin{tabular}[c]{@{}c@{}}solver,\\ shrinkage\end{tabular}                                         & \begin{tabular}[c]{@{}c@{}}BOs,\\ Hyperband\end{tabular}                    & \begin{tabular}[c]{@{}c@{}}Skpot,\\ Hyperopt,\\ SMAC,\\ Hyperband\end{tabular}                      \\ \Xhline{1.2pt}

\end{tabular}
\label{t2o}%
\end{table*}

To put theory into practice, several experiments have been conducted based on Table 2. This section provides the experiments of applying eight different HPO techniques to three common and representative ML algorithms on two benchmark datasets. In the first part of this section, the experimental setup and the main process of HPO are discussed. In the second part, the results of utilizing different HPO methods are compared and analyzed. The sample code of the experiments has been published in \cite{github} to illustrate the process of applying hyper-parameter optimization to ML models. 

\subsection{Experimental Setup}
Based on the steps to optimize hyper-parameters discussed in Section 2.2, several steps were completed before the actual optimization experiments start. 

Firstly, two standard benchmarking datasets provided by the sklearn library \cite{sklearn}, namely, the Modified National Institute of Standards and Technology dataset (MNIST) and the Boston housing dataset, are selected as the benchmark datasets for HPO method evaluation on data analytics problems.  MNIST is a hand-written digit recognition dataset used as a multi-classification problem, while the Boston housing dataset contains information about the price of houses in various places in the city of Boston and can be used as a regression dataset to predict the housing prices. 

At the next stage, the ML models with their objective function need to be configured. In Section 5.2, all common ML models are divided into five categories based on their hyper-parameter types. Among those ML categories, "one discrete hyper-parameter", "a few conditional hyper-parameters", and "a large hyper-parameter configuration space with multiple types of hyper-parameters" are the three most common cases. Thus, three ML algorithms, KNN, SVM, and RF, are selected as the target models to be optimized, since their hyper-parameter types represent the three most common HPO cases: KNN has one important hyper-parameter, the number of considered nearest neighbors for each sample; SVM has a few conditional hyper-parameters, like the kernel type and the penalty parameter $C$; RF has multiple hyper-parameters of different types, as discussed in Section 3.1.5. Moreover, KNN, SVM, and RF can all be applied to solve both classification and regression problems. 

In the next step, the performance metrics and evaluation methods are configured. For each experiment on the selected two datasets, 3-fold cross validation is implemented to evaluate the involved HPO methods. The two most commonly-used performance metrics are used in our experiments. For classification models, accuracy is used as the classifier performance metric, which is the proportion of correctly classified data; while for regression models, the mean squared error (MSE) is used as the regressor performance metric, which measures the average squared difference between the predicted values and the actual values. Additionally, the computational time (CT) , the total time needed to complete a HPO process with 3-fold cross-validation, is also used as the model efficiency metric \cite{IDSme}.
In each experiment, the optimized ML model architecture that has the highest accuracy or the lowest MSE and the optimal hyper-parameter configuration will be returned.

After that, to fairly compare different optimization algorithms and frameworks, certain constraints should be satisfied. Firstly, we compare different HPO methods using the same hyper-parameter configuration space.  For KNN, the only hyper-parameter to be optimized, 'n\_neighbors', is set to be in the same range of 1 to 20 for each optimization method evaluation. The hyper-parameters of SVM and RF models for classification and regression problems are also set to be in the same configuration space for each type of problem.  The specifics of the configuration space for ML models are shown in Table \ref{ta1}. The selected hyper-parameters and their search space are determined based on the concepts in Section 3, domain knowledge, and manual testings \cite{grid1}. The hyper-parameter types of each ML algorithm are also summarized in Table \ref{ta1}.

\begin{table}
\centering
\caption{Configuration space for the hyper-parameters of tested ML models}
\setlength\extrarowheight{1pt}
\centering
\scriptsize
\begin{tabular}{l|l|l|l} 
\Xhline{1.2pt}
\textbf{ML Model }                       & \multicolumn{1}{l|}{\textbf{Hyper-parameter}} & \multicolumn{1}{l|}{\textbf{Type}}        & \textbf{Search Space}                            \\ 
\Xhline{1.2pt}
\multirow{6}{*}{RF Classifier}  & n\_estimators                        & Discrete                         & {[}10,100]                              \\ 
\cline{2-4}
                                & max\_depth                           & \multicolumn{1}{l|}{Discrete}    & {[}5,50]                                \\ 
\cline{2-4}
                                & min\_samples\_split                  & Discrete                         & {[}2,11]                                \\ 
\cline{2-4}
                                & min\_samples\_leaf                   & Discrete                         & {[}1,11]                                \\ 
\cline{2-4}
                                & criterion                            & \multicolumn{1}{l|}{Categorical} & {[}'gini', 'entropy']                   \\ 
\cline{2-4}
                                & max\_features                        & Discrete                         & {[}1,64]                                \\ 
\hline
\multirow{2}{*}{SVM Classifier} & C                                    & Continuous                       & {[}0.1,50]                              \\ 
\cline{2-4}
                                & kernel                               & Categorical                      & {[}'linear', 'poly', 'rbf', 'sigmoid']  \\ 
\hline
KNN Classifier                  & n\_neighbors                         & Discrete                         & {[}1,20]                                \\ 
\hline
\multirow{6}{*}{RF Regressor}   & n\_estimators                        & Discrete                         & {[}10,100]                              \\ 
\cline{2-4}
                                & max\_depth                           & Discrete                         & {[}5,50]                                \\ 
\cline{2-4}
                                & min\_samples\_split                  & Discrete                         & {[}2,11]                                \\ 
\cline{2-4}
                                & min\_samples\_leaf                   & Discrete                         & {[}1,11]                                \\ 
\cline{2-3}\cline{4-4}
                                & criterion                            & Categorical                      & {[}'mse', 'mae']                        \\ 
\cline{2-4}
                                & max\_features                        & Discrete                         & {[}1,13]                                \\ 
\hline
\multirow{3}{*}{SVM Regressor}  & C                                    & Continuous                       & {[}0.1,50]                              \\ 
\cline{2-4}
                                & kernel                               & Categorical                      & {[}'linear', 'poly', 'rbf', 'sigmoid']  \\ 
\cline{2-4}
                                & epsilon                              & Continuous                       & {[}0.001,1]                             \\ 
\hline
KNN Regressor                   & n\_neighbors                         & Discrete                         & {[}1,20]                                \\
\Xhline{1.2pt}
\end{tabular}
\label{ta1}%
\end{table}

On the other hand, to fairly compare the performance metrics of optimization techniques, the maximum number of iterations for all HPO methods is set to 50 for RF and SVM model optimizations, and 10 for KNN model optimization based on manual testings and domain knowledge. 
Moreover, to avoid the impacts of randomness, all experiments are repeated ten times with different random seeds, and results are averaged for regression problems or given the majority vote for classification problems.

In Section 4, more than ten HPO methods are introduced. In our experiments, eight representative HPO approaches are selected for performance comparison, including GS, RS, BO-GP, BO-TPE, Hyperband, BOHB, GA, and PSO. After setting up the fair experimental environments for each HPO method, the HPO experiments are implemented based on the steps discussed in Section 2.2.

All experiments were conducted using Python 3.5 on a machine with 6 Core i7-8700 processor and 16 gigabytes (GB) of memory. The involved ML and HPO algorithms are evaluated using multiple open-source Python libraries and frameworks introduced in Section 6, including sklearn \cite{sklearn}, Skopt \cite{SKOPT}, Hyperopt \cite{Hyperopt}, Optunity \cite{Optunity}, Hyperband \cite{Hyperband}, BOHB \cite{BOHB}, and TPOT \cite{TPOT}. 

\subsection{Performance Comparison}
The experiments of applying eight different HPO methods to ML models are summarized in Tables \ref{te1} to \ref{te6}. Tables \ref{te1} to \ref{te3} provide the performance of each optimization algorithm when applied to RF, SVM, and KNN classifiers evaluated on the MNIST dataset after a complete optimization process; while Tables \ref{te4} to \ref{te6} demonstrate the performance of each HPO method when applied to RF, SVM, and KNN regressors evaluated on the Boston-housing dataset. In the first step, each ML model with its default hyper-parameter configuration is trained and evaluated as the baseline model. After that, each HPO algorithm is implemented on the ML models to evaluate and compare their accuracies for classification problems, or MSEs for regression problems, as well as their computational time (CT). 

\begin{table}
\centering
\caption{Performance evaluation of applying HPO methods to the RF classifier on the MNIST dataset}
\setlength\extrarowheight{1pt}
\centering
\scriptsize
\begin{tabular}{p{2cm}|p{1.5cm}|p{1.5cm}} 
\Xhline{1.2pt}
\textbf{Optimization Algorithm} & \textbf{Accuracy} (\%) & \textbf{CT} (s)  \\ 
\Xhline{1.2pt}
Default HPs            & 90.65         & 0.09    \\ 
\hline
GS                     & 93.32         & 48.62   \\ 
\hline
RS                     & 93.38         & 16.73   \\ 
\hline
BO-GP                  & 93.38         & 20.60   \\ 
\hline
BO-TPE                 & 93.88         & 12.58   \\ 
\hline
Hyperband              & 93.38         & 8.89    \\ 
\hline
BOHB                   & 93.38         & 9.45    \\ 
\hline
GA                     & 93.83         & 19.19   \\ 
\hline
PSO                    & 93.73         & 12.43   \\
\Xhline{1.2pt}
\end{tabular}
\label{te1}%
\end{table}

\begin{table}
\centering
\caption{Performance evaluation of applying HPO methods to the SVM classifier on the MNIST dataset}
\setlength\extrarowheight{1pt}
\centering
\scriptsize
\begin{tabular}{p{2cm}|p{1.5cm}|p{1.5cm}} 
\Xhline{1.2pt}
\textbf{Optimization Algorithm} & \textbf{Accuracy} (\%) & \textbf{CT} (s)  \\ 
\Xhline{1.2pt}
Default HPs & 97.05 & 0.29   \\ 
\hline
GS          & 97.44 & 32.90  \\ 
\hline
RS          & 97.35 & 12.48  \\ 
\hline
BO-GP       & 97.50 & 17.56  \\ 
\hline
BO-TPE      & 97.44 & 3.02   \\ 
\hline
Hyperband   & 97.44 & 11.37  \\ 
\hline
BOHB        & 97.44 & 8.18   \\ 
\hline
GA          & 97.44 & 16.89  \\ 
\hline
PSO         & 97.44 & 8.33   \\
\Xhline{1.2pt}
\end{tabular}
\label{te2}%
\end{table}

\begin{table}
\centering
\caption{Performance evaluation of applying HPO methods to the KNN classifier on the MNIST dataset}
\setlength\extrarowheight{1pt}
\centering
\scriptsize
\begin{tabular}{p{2cm}|p{1.5cm}|p{1.5cm}}  
\Xhline{1.2pt}
\textbf{Optimization Algorithm} & \textbf{Accuracy} (\%) & \textbf{CT} (s)  \\ 
\Xhline{1.2pt}
Default HPs & 96.27 & 0.24  \\ 
\hline
GS          & 96.22 & 7.86  \\ 
\hline
RS          & 96.33 & 6.44  \\ 
\hline
BO-GP       & 96.83 & 1.12  \\ 
\hline
BO-TPE      & 96.83 & 2.33  \\ 
\hline
Hyperband   & 96.22 & 4.54  \\ 
\hline
BOHB        & 97.44 & 3.84  \\ 
\hline
GA          & 96.83 & 2.34  \\ 
\hline
PSO         & 96.83 & 1.73  \\
\hline
\Xhline{1.2pt}
\end{tabular}
\label{te3}%
\end{table}

\begin{table}
\centering
\caption{Performance evaluation of applying HPO methods to the RF regressor on the Boston-housing dataset}
\setlength\extrarowheight{1pt}
\centering
\scriptsize
\begin{tabular}{p{2cm}|p{1.5cm}|p{1.5cm}} 
\Xhline{1.2pt}
\textbf{Optimization Algorithm} & \textbf{MSE} & \textbf{CT} (s)  \\ 
\Xhline{1.2pt}
Default HPs & 31.26 & 0.08   \\ 
\hline
GS          & 29.02 & 4.64   \\ 
\hline
RS          & 27.92 & 3.42   \\ 
\hline
BO-GP       & 26.79 & 17.94  \\ 
\hline
BO-TPE      & 25.42 & 1.53   \\ 
\hline
Hyperband   & 26.14 & 2.56   \\ 
\hline
BOHB        & 25.56 & 1.88   \\ 
\hline
GA          & 26.95 & 4.73   \\ 
\hline
PSO         & 25.69 & 3.20   \\
\Xhline{1.2pt}
\end{tabular}
\label{te4}%
\end{table}

\begin{table}
\centering
\caption{Performance evaluation of applying HPO methods to the SVM regressor on the Boston-housing dataset}
\setlength\extrarowheight{1pt}
\centering
\scriptsize
\begin{tabular}{p{2cm}|p{1.5cm}|p{1.5cm}} 
\Xhline{1.2pt}
\textbf{Optimization Algorithm} & \textbf{MSE} & \textbf{CT} (s)  \\ 
\Xhline{1.2pt}
Default HPs            & 77.43 & 0.02    \\ 
\hline
GS                     & 67.07 & 1.33    \\ 
\hline
RS                     & 61.40 & 0.48    \\ 
\hline
BO-GP                  & 61.27 & 5.87    \\ 
\hline
BO-TPE                 & 59.40 & 0.33    \\ 
\hline
Hyperband              & 73.44 & 0.32    \\ 
\hline
BOHB                   & 59.67 & 0.31    \\ 
\hline
GA                     & 60.17 & 1.12    \\ 
\hline
PSO                    & 58.72 & 0.53    \\
\Xhline{1.2pt}
\end{tabular}
\label{te5}%
\end{table}

\begin{table}
\centering
\caption{Performance evaluation of applying HPO methods to the KNN regressor on the Boston-housing dataset}
\setlength\extrarowheight{1pt}
\centering
\scriptsize
\begin{tabular}{p{2cm}|p{1.5cm}|p{1.5cm}} 
\Xhline{1.2pt}
\textbf{Optimization Algorithm} & \textbf{MSE} & \textbf{CT} (s)  \\ 
\Xhline{1.2pt}
Default HPs & 81.48 & 0.004  \\ 
\hline
GS          & 81.53 & 0.12   \\ 
\hline
RS          & 80.77 & 0.11   \\ 
\hline
BO-GP       & 80.77 & 0.49   \\ 
\hline
BO-TPE      & 80.83 & 0.08   \\ 
\hline
Hyperband   & 80.87 & 0.10   \\ 
\hline
BOHB        & 80.77 & 0.09   \\ 
\hline
GA          & 80.77 & 0.33   \\ 
\hline
PSO         & 80.74 & 0.19   \\
\Xhline{1.2pt}
\end{tabular}
\label{te6}%
\end{table}

From Tables \ref{te1} to \ref{te6}, we can see that using the default HP configurations do not yield the best model performance in our experiments, which emphasizes the importance of utilizing HPO methods. GS and RS can be seen as baseline models for HPO problems. From the results in Tables \ref{te1} to \ref{te6}, it is shown that the computational time of GS is often much higher than other optimization methods. With the same search space size, RS is faster than GS, but both of them cannot guarantee to detect the near-optimal hyper-parameter configurations of ML models, especially for RF and SVM models which have a larger search space than KNN.

The performance of BO and multi-fidelity models is much better than GS and RS. The computation time of BO-GP is often higher than other HPO methods due to its cubic time complexity, but it can obtain better performance metrics for ML models with small-size continuous hyper-parameter space, like KNN. Conversely, hyperband is often not able to obtain the highest accuracy or the lowest MSE among the optimization methods, but their computational time is low because it works on the small-sized subsets. The performance of BO-TPE and BOHB is often better than others, since they can detect the optimal or near-optimal hyper-parameter configurations within a short computational time. 

For metaheuristics methods, GA and PSO, their accuracies are often higher than other HPO methods for classification problems, and their MSEs are often lower than other optimization techniques. However, their computational time is often higher than BO-TPE and multi-fidelity models, especially for GA, which does not support parallel executions. 

To summarize, it is simple to implement GS and RS, but they often cannot detect the optimal hyper-parameter configurations or cost much computational time. BO-GP and GA also cost more computational time than many other HPO methods, but BO-GP works well on small configuration space, while GA is effective for large configuration space. Hyperband's computational time is low, but it cannot guarantee to detect the global optimums. For ML models with large configuration space, BO-TPE, BOHB, and PSO often work well. 

\section{Open Issues, Challenges, and Future Research Directions}
Although there have been many existing HPO algorithms and practical frameworks, some issues still need to be addressed, and several aspects in this domain could be improved. In this section, we discuss the open challenges, current research questions, and potential research directions in the future. They can be classified as model complexity challenges and model performance challenges, as summarized in Table \ref{t3}.

\begin{table*}[ht]
\caption{The open challenges and future directions of HPO research}
\setlength\extrarowheight{1pt}
\centering
\scriptsize
\begin{tabular}{l|p{4cm}|p{6cm}}
\Xhline{1.2pt}
\textbf{Category}                                        & \textbf{Challenges \& Future Requirements} & \textbf{Brief Description}                                                                                                                                             \\ \Xhline{1.2pt}
\multirow{2}{*}{Model complexity}                        & Costly objective function evaluations      & {HPO methods should reduce evaluation time on large datasets.}                                                                                       \\ \cline{2-3} 
                                                         & Complex search space                       & HPO methods should reduce execution time on high dimensionalities (large hyper-parameter search space). \\ \hline
\multicolumn{1}{l|}{\multirow{7}{*}{Model performance}} & Strong anytime performance                 & HPO methods should be able to detect the optimal or near-optimal HPs even with a very limited budget.             \\ \cline{2-3} 
\multicolumn{1}{l|}{}                                   & Strong final performance                   & {HPO methods should be able to detect the global optimum when given a sufficient budget.}                                                                       \\ \cline{2-3} 
\multicolumn{1}{l|}{}                                   & Comparability                              & There should exist a standard set of benchmarks to fairly evaluate and compare different optimization algorithms.   \\ \cline{2-3} 
\multicolumn{1}{l|}{}                                   & Over-fitting and generalization            & The optimal HPs detected by HPO methods should have generalizability to build efficient models on unseen data.                                                             \\ \cline{2-3} 
\multicolumn{1}{l|}{}                                   & Randomness                                 & HPO methods should reduce randomness on the obtained results.                                                                                      \\ \cline{2-3} 
\multicolumn{1}{l|}{}                                   & Scalability                                & HPO methods should be scalable to multiple libraries or platforms (\textit{e.g.}, distributed ML platforms).     \\ \cline{2-3} 
\multicolumn{1}{l|}{}                                   & Continuous updating capability             & HPO methods should consider their capacity to detect and update optimal HP combinations on continuously-updated data.  \\ \Xhline{1.2pt}

\end{tabular}
\label{t3}%
\end{table*}

\subsection{Model Complexity}
\subsubsection{Costly Objective Function Evaluations}
To evaluate the performance of a ML model with different hyper-parameter configurations, its objective function must be minimized in each evaluation. Depending on the scale of data, the model complexity, and available computational resources, the evaluation of each hyper-parameter configuration may cost several minutes, hours, days, or even more \cite{HPS}. Additionally, the values of certain hyper-parameters have a direct impact on the execution time, like the number of considered neighbors in KNN, the number of basic decision trees in RF, and the number of hidden layers in deep neural networks \cite{NNetime}. 

To solve this problem by HPO algorithms, BO models reduce the total number of evaluations by spending time choosing the next evaluating point instead of simply evaluating all possible hyper-parameter configurations; however, they still require much execution time due to their poor capacity for parallelization. On the other hand, although multi-fidelity optimization methods, like Hyperband, have had some success dealing with HPO problems with limited budgets, there are still some problems that cannot be effectively solved by HPO due to the complexity of models or the scale of datasets \cite{AMLB}. For example, the ImageNet \cite{Imagenet} challenge is a very popular problem in the image processing domain, but there has not been any research or work on efficiently optimizing hyper-parameters for the ImageNet challenge yet, due to its huge scale and the complexity of CNN models used on ImageNet.
\subsubsection{Complex Search Space}
In many problems to which ML algorithms are applied, only a few hyper-parameters have significant effects on model performance, and they are the main hyper-parameters that require tuning. However, certain other unimportant hyper-parameters may still affect the performance slightly and may be considered to optimize the ML model further, which increases the dimensionality of hyper-parameter search space. As the number of hyper-parameters and configurations increase, they exponentially increase the dimensionality of the search space and the complexity of the problems, and the total objective function evaluation time will also increase exponentially \cite{EHPO}. Therefore, it is necessary to reduce the influence of large search spaces on execution time by improving existing HPO methods.
\subsection{Model Performance}
\subsubsection{Strong Anytime Performance and Final Performance}
HPO techniques are often expensive and sometimes require extreme resources, especially for massive datasets or complex ML models. One example of a resource-intensive model is deep learning models, since they view objective function evaluations as black-box functions and do not consider their complexity. However, the overall budget is often very limited for most practical situations, so HPO algorithms should be able to prioritize objective function evaluations and have a strong anytime performance, which indicates the capacity to detect optimal or near-optimal configurations even with a very limited budget \cite{BOHB}. For instance, an efficient HPO method should have a high convergence speed so that there would not be a huge difference between the results before and after model convergence, and should avoid random results even if time and resources are limited, like RS methods cannot. 

On the other hand, if conditions permit and an adequate budget is given, HPO approaches should be able to identify the global optimal hyper-parameter configuration, named a strong final performance \cite{BOHB}.
\subsubsection{Comparability of HPO Methods}
To optimize the hyper-parameters of ML models, different optimization algorithms can be applied to each ML framework. Different optimization techniques have their own strengths and drawbacks in different cases, and currently, there is no single optimization approach that outperforms all other approaches when processing different datasets with various metrics and hyper-parameter types \cite{AMLSC}. In this paper, we have analyzed the strengths and weaknesses of common hyper-parameter optimization techniques based on their principles and their performance in practical applications; but this topic could be extended more comprehensively.

To solve this problem, a standard set of benchmarks could be designed and agreed on by the community for a better comparison of different HPO algorithms. For example, there is a platform called COCO (Comparing Continuous Optimizers) \cite{COCO} that provides benchmarks and analyzes common continuous optimizers. However, there is, to date, not any reliable platform that provides benchmarks and analysis of all common hyper-parameter optimization approaches. It would be easier for people to choose HPO algorithms in practical applications if a platform like COCO exists for HPO problems. In addition, a unified metric can also improve the comparability of different HPO algorithms, since different metrics are currently used in different practical problems \cite{AMLB}.

On the other hand, based on the comparison of different HPO algorithms, a way to further improve HPO is to combine existing models or propose new models that contain as many benefits as possible and are more suitable for practical problems than existing singular models. For example, the BOHB method \cite{BOHB} has had some success dealing with HPO problems by combining Bayesian optimization and Hyperband. In addition, future research should consider both model performance and time budgets to develop HPO algorithms that suit real-world applications.
\subsubsection{Over-fitting and Generalization}
Generalization is another issue with HPO models. Since hyper-parameter evaluations are done with a finite number of evaluations in datasets, the optimal hyper-parameter values detected by HPO approaches might not be the same optimums on previously-unseen data. This is similar to over-fitting issues with ML models that occur when a model is closely fit to a finite number of known data points but is unfit to unseen data \cite{over-fitting}. Generalization is also a common concern for multi-fidelity algorithms, like Hyperband and BOHB, since they need to extract subsets to represent the entire dataset.

One solution to reduce or avoid over-fitting is to use cross-validation to identify a stable optimum that performs best in all or most of the subsets instead of a sharp optimum that only performs well in a singular validation set \cite{AMLB}. However, cross-validation increases the execution time several-fold. It would be beneficial if methods can better deal with overfitting and improve generalization in future research.
\subsubsection{Randomness}
There are stochastic components in the objective function of ML algorithms; thus, in some cases, the optimal hyper-parameter configuration might be different after each run. This randomness could be due to various procedures of certain ML models, like neural network initialization, or different sampled subsets in a bagging model \cite{HPS}; or due to certain procedures of HPO algorithms, like crossover and mutation operations in GA. In addition, it is often difficult for HPO methods to identify the global optimums, due to the fact that HPO problems are mainly NP-hard problems. Many existing HPO algorithms can only collect several different near-optimal values, which is caused by randomness. Thus, the existing HPO models can be further improved to reduce the impact of randomness. One possible solution is to run a HPO method multiple times and select the hyper-parameter value that occurs most as the final optimum.
\subsubsection{Scalability}
In practice, one main limitation of many existing HPO frameworks is that they are tightly integrated with one or a couple of machine learning libraries, like sklearn and Keras, which restricts them to only work with a single node instead of large data volumes \cite{AMLSC}. To tackle large datasets, some distributed machine learning platforms, like Apache SystemML \cite{SystemML} and Spark MLib \cite{MLib}, have been developed; however, only very few HPO frameworks exist that support distributed ML. Therefore, more research efforts and scalable HPO frameworks, like the ones supporting distributed ML platforms, should be developed to support more libraries.

On the other hand, future practical HPO algorithms should have the scalability to efficiently optimize hyper-parameters from a small size to a large size, irrespective of whether they are continuous, discrete, categorical, or conditional hyper-parameters.
\subsubsection{Continuous Updating Capability}
In practice, many datasets are not stationary and are constantly updated by adding new data and deleting old data. Correspondingly, the optimal hyper-parameter values or combinations may also change with the changes in data. Currently, developing HPO methods with the capacity to continuously tune hyper-parameter values as the data changes has not drawn much attention, since researchers and data analysts often do not alter the ML model after achieving a currently optimal performance \cite{AMLSC}. However, since their optimal hyper-parameter values would change as data changes, proper approaches should be proposed to achieve continuous updating capability.
\section{Conclusion}
Machine learning has become the primary strategy for tackling data-related problems and has been widely used in various applications. To apply ML models to practical problems, their hyper-parameters need to be tuned to fit specific datasets. However, since the scale of produced data is greatly increased in real-life, and manually tuning hyper-parameters is extremely computationally expensive, it has become crucial to optimize hyper-parameters by an automatic process. In this survey paper, we have comprehensively discussed the state-of-the-art research into the domain of hyper-parameter optimization as well as how to apply them to different ML models by theory and practical experiments. To apply optimization methods to ML models, the hyper-parameter types in a ML model is the main concern for HPO method selection. To summarize, BOHB is the recommended choice for optimizing a ML model, if randomly selected subsets are highly-representative of the given dataset, since it can efficiently optimize all types of hyper-parameters; otherwise, BO models are recommended for small hyper-parameter configuration space, while PSO is usually the best choice for large configuration space. 
Moreover, some existing useful HPO tools and frameworks, open challenges, and potential research directions are also provided and highlighted for practical use and future research purposes. We hope that our survey paper serves as a useful resource for ML users, developers, data analysts, and researchers to use and tune ML models utilizing proper HPO techniques and frameworks. We also hope that it helps to enhance understanding of the challenges that still exist within the HPO domain, and thereby further advancing HPO and ML applications in future research.

\newpage
\begin{wrapfigure}{l}{30mm} 
\includegraphics[width=1.25in,height=1.5in,clip,keepaspectratio]{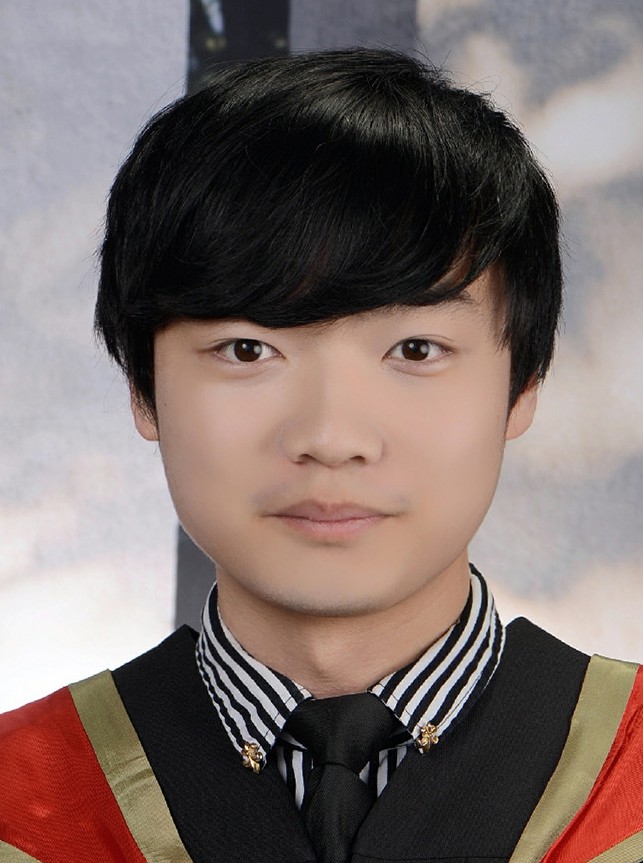}
\end{wrapfigure}\par
\textbf{Li Yang} received his Ph.D. in Electrical and Computer Engineering from Western University, London, Canada, in August 2022, his MASc. degree in Engineering from the University of Guelph, Guelph, Canada, 2018, and his B.E. degree in Computer Science from Wuhan University of Science and Technology, Wuhan, China, in 2016. Currently, he is a Postdoctoral Associate and Sessional Lecturer in the Optimized Computing and Communications (OC2) Lab at Western University.  His research interests include cybersecurity, machine learning, AutoML, deep learning, network data analytics, Internet of Things (IoT), anomaly detection, online learning, concept drift, and time series data analytics.\par
  
~\\

\begin{wrapfigure}{l}{30mm} 
\includegraphics[width=1.25in,height=1.45in,clip,keepaspectratio]{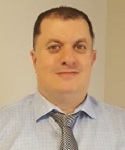}
\end{wrapfigure}\par
\textbf{Abdallah Shami} is currently a Professor in the Electrical and Computer Engineering Department and the Acting Associate Dean (Research) of the Faculty of Engineering, Western University, London, ON, Canada, where he is also the Director of the Optimized Computing and Communications Laboratory. Dr. Shami has chaired key symposia for the IEEE GLOBECOM, IEEE International Conference on Communications, and IEEE International Conference on Computing, Networking and Communications. He was the elected Chair for the IEEE Communications Society Technical Committee on Communications Software from 2016 to 2017 and the IEEE London Ontario Section Chair from 2016 to 2018. He is currently an Associate Editor of the IEEE Transactions on Mobile Computing, IEEE Internet of Things, and IEEE Communications Surveys and Tutorials journals.\par

\begin{thebibliography}{00}

\bibitem{MLsc} M.I. Jordan, T.M. Mitchell, Machine learning: Trends, perspectives, and prospects, Science      349 (2015) 255–260. https://doi.org/10.1126/science.aaa8415.
\bibitem{SOAML} M.-A. Zöller and M. F. Huber,  Benchmark and Survey of Automated Machine Learning Frameworks,  arXiv preprint arXiv:1904.12054, (2019). https://arxiv.org/abs/1904.12054.
\bibitem{AMLSC} R. E. Shawi, M. Maher, S. Sakr, Automated machine learning:  State-of-the-art and open challenges, arXiv preprint arXiv:1906.02287, (2019). http://arxiv.org/abs/1906.02287.
\bibitem{2Ps}	M. Kuhn and K. Johnson, Applied Predictive Modeling, Springer (2013) ISBN: 9781461468493.
\bibitem{parameters} G.I. Diaz, A. Fokoue-Nkoutche, G. Nannicini, H. Samulowitz, An effective algorithm for hyperparameter optimization of neural networks, IBM J. Res. Dev. 61 (2017) 1–20. https://doi.org/10.1147/JRD.2017.2709578.
\bibitem{AMLB}	F. Hutter, L. Kotthoff, and J. Vanschoren, Eds., Automatic Machine Learning: Methods, Systems, Challenges, Springer (2019) ISBN: 9783030053185.
\bibitem{EHPO}	N. Decastro-García, Á. L. Muñoz Castañeda, D. Escudero García, and M. V. Carriegos, Effect of the Sampling of a Dataset in the Hyperparameter Optimization Phase over the Efficiency of a Machine Learning Algorithm, Complexity 2019 (2019). https://doi.org/10.1155/2019/6278908.
\bibitem{ADL} S. Abreu, Automated Architecture Design for Deep Neural Networks, arXiv preprint arXiv:1908.10714, (2019). http://arxiv.org/abs/1908.10714.
\bibitem{BBHPO}	O. S. Steinholtz, A Comparative Study of Black-box Optimization Algorithms for Tuning of Hyper-parameters in Deep Neural Networks, M.S. thesis, Dept. Elect. Eng., Luleå Univ. Technol., (2018).
\bibitem{ASHPO}	G. Luo, A review of automatic selection methods for machine learning algorithms and hyper-parameter values, Netw. Model. Anal. Heal. Informatics Bioinforma. 5 (2016) 1–16. https://doi.org/10.1007/s13721-016-0125-6.
\bibitem{GBad}	D. Maclaurin, D. Duvenaud, R.P. Adams, Gradient-based Hyperparameter Optimization through Reversible Learning, arXiv preprint arXiv:1502.03492, (2015). http://arxiv.org/abs/1502.03492.
\bibitem{AHPO} J. Bergstra, R. Bardenet, Y. Bengio, and B. K\'{e}gl, Algorithms for hyper-parameter optimization, Proc. Adv. Neural Inf. Process. Syst., (2011) 2546–2554.
\bibitem{RS} B. James and B. Yoshua, Random Search for Hyper-Parameter Optimization, J. Mach. Learn. Res. 13 (1) (2012) 281–305.
\bibitem{BOHP} K. Eggensperger, M. Feurer, F. Hutter, J. Bergstra, J. Snoek, H. Hoos, K. Leyton-Brown, Towards an Empirical Foundation for Assessing Bayesian Optimization of Hyperparameters, BayesOpt Work. (2013) 1–5.
\bibitem{surrogates} K. Eggensperger, F. Hutter, H.H. Hoos, K. Leyton-Brown, Efficient benchmarking of hyperparameter optimizers via surrogates, Proc. Natl. Conf. Artif. Intell. 2 (2015) 1114–1120.
\bibitem{Hyperband}	L. Li, K. Jamieson, G. DeSalvo, A. Rostamizadeh, and A. Talwalkar, Hyperband: A novel bandit-based approach to hyperparameter optimization, J. Mach. Learn. Res. 18 (2012) 1–52.
\bibitem{HumanAML}	Q. Yao \textit{et al.}, Taking Human out of Learning Applications: A Survey on Automated Machine Learning, arXiv preprint arXiv:1810.13306, (2018). http://arxiv.org/abs/1810.13306.
\bibitem{GA2}	S. Lessmann, R. Stahlbock, S.F. Crone, Optimizing hyperparameters of support vector machines by genetic algorithms, Proc. 2005 Int. Conf. Artif. Intell. ICAI’05. 1 (2005) 74–80.
\bibitem{PSODL}	P. R. Lorenzo, J. Nalepa, M. Kawulok, L. S. Ramos, and J. R. Paster, Particle swarm optimization for hyper-parameter selection in deep neural networks, Proc. ACM Int. Conf. Genet. Evol. Comput., (2017) 481–488.
\bibitem{mathopt1}	S. Sun, Z. Cao, H. Zhu, J. Zhao, A Survey of Optimization Methods from a Machine Learning Perspective, arXiv preprint arXiv:1906.06821, (2019). https://arxiv.org/abs/1906.06821.
\bibitem{mathopt2}	T.M. S. Bradley, A. Hax, Applied Mathematical Programming, Addison-Wesley, Reading, Massachusetts. (1977).
\bibitem{convex}	S. Bubeck, Convex optimization: Algorithms and complexity, Found. Trends Mach. Learn. 8 (2015) 231–357. https://doi.org/10.1561/2200000050.
\bibitem{UBO}	B. Shahriari, A. Bouchard-Côté, and N. de Freitas, “Unbounded Bayesian optimization via regularization,” Proc. Artif. Intell. Statist., (2016) 1168–1176.
\bibitem{HPOpro} G.I. Diaz, A. Fokoue-Nkoutche, G. Nannicini, H. Samulowitz, An effective algorithm for hyperparameter optimization of neural networks, IBM J. Res. Dev. 61 (2017) 1–20. https://doi.org/10.1147/JRD.2017.2709578.
\bibitem{OML}	C. Gambella, B. Ghaddar, and J. Naoum-Sawaya, Optimization Models for Machine Learning: A Survey, arXiv preprint arXiv:1901.05331, (2019). http://arxiv.org/abs/1901.05331. 
\bibitem{AUTOMS} E. R. Sparks, A. Talwalkar, D. Haas, M. J. Franklin, M. I. Jordan, and T. Kraska, Automating model search for large scale machine learning, Proc. 6th ACM Symp. Cloud Comput., (2015) 368–380.
\bibitem{NumericalO}	 J. Nocedal and S. Wright, Numerical Optimization, (2006) Springer-Verlag, ISBN: 978-0-387-40065-5.
\bibitem{superAM}	 A. Moubayed, M. Injadat, A. Shami, H. Lutfiyya, DNS Typo-Squatting Domain Detection: A Data Analytics \& Machine Learning Based Approach, 2018 IEEE Glob. Commun. Conf. GLOBECOM 2018 - Proc. (2018). https://doi.org/10.1109/GLOCOM.2018.8647679.
\bibitem{supervised} R. Caruana, A. Niculescu-Mizil, An empirical comparison of supervised learning algorithms, ACM Int. Conf. Proceeding Ser. 148 (2006) 161–168. https://doi.org/10.1145/1143844.1143865.
\bibitem{sklearnbook} O. Kramer, Scikit-Learn, in Machine Learning for Evolution Strategies. Cham, Switzerland: Springer International Publishing, (2016) 45–53.
\bibitem{sklearn} F. Pedregosa \textit{et al.}, Scikit-learn: Machine learning in Python, J. Mach. Learn. Res., 12 (2011) 2825–2830.
\bibitem{xgboost} T.Chen, C.Guestrin, XGBoost: a scalable tree boosting system, arXiv preprint arXiv:1603.02754, (2016). http://arxiv.org/abs/1603.02754. 
\bibitem{keras}	F. Chollet, Keras, 2015. https://github.com/fchollet/keras.
\bibitem{SL} C. Gambella, B. Ghaddar, J. Naoum-Sawaya, Optimization Models for Machine Learning: A Survey, (2019) 1–40. http://arxiv.org/abs/1901.05331
\bibitem{ML}	C.M. Bishop, Pattern Recognition and Machine Learning. (2006) Springer, ISBN: 978-0-387-31073-2.
\bibitem{ridge}	A.E. Hoerl, R.W. Kennard, Ridge Regression: Applications to Nonorthogonal Problems, Technometrics. 12 (1970) 69–82. https://doi.org/10.1080/00401706.1970.10488635.
\bibitem{ridgelasso} L.E. Melkumova, S.Y. Shatskikh, Comparing Ridge and LASSO estimators for data analysis, Procedia Eng. 201 (2017) 746–755. https://doi.org/10.1016/j.proeng.2017.09.615.
\bibitem{lasso}	R. Tibshirani, Regression Shrinkage and Selection Via the Lasso, J. R. Stat. Soc. Ser. B. 58 (1996) 267–288. https://doi.org/10.1111/j.2517-6161.1996.tb02080.x.
\bibitem{LoR}	D.W. Hosmer Jr, S. Lemeshow, Applied logistic regression, Technometrics, 34 (1) (2013), 358-359.
\bibitem{LRnorms} J.O. Ogutu, T. Schulz-Streeck, H.P. Piepho, Genomic selection using regularized linear regression models: ridge regression, lasso, elastic net and their extensions, BMC Proceedings. BioMed Cent. 6 (2012).
\bibitem{KNN}	J.M. Keller, M.R. Gray, A Fuzzy K-Nearest Neighbor Algorithm, IEEE Trans. Syst. Man Cybern. SMC-15 (1985) 580–585. https://doi.org/10.1109/TSMC.1985.6313426.
\bibitem{kinknn} W. Zuo, D. Zhang, K. Wang, On kernel difference-weighted k-nearest neighbor classification, Pattern Anal. Appl. 11 (2008) 247–257. https://doi.org/10.1007/s10044-007-0100-z.
\bibitem{SVM1}	A. Smola, V. Vapnik, Support vector regression machines, Adv. Neural Inf. Process. Syst. 9 (1997) 155-161.
\bibitem{SVMme} L. Yang, R. Muresan, A. Al-Dweik, L.J. Hadjileontiadis, Image-Based Visibility Estimation Algorithm for Intelligent Transportation Systems, IEEE Access. 6 (2018) 76728–76740. https://doi.org/10.1109/ACCESS.2018.2884225.
\bibitem{SVMme2}	L. Yang, Comprehensive Visibility Indicator Algorithm for Adaptable Speed Limit Control in Intelligent Transportation Systems, M.A.Sc. thesis, University of Guelph, 2018.
\bibitem{SVMkernel} O.S. Soliman, A.S. Mahmoud, A classification system for remote sensing satellite images using support vector machine with non-linear kernel functions, 2012 8th Int. Conf. Informatics Syst. INFOS 2012. (2012) BIO-181-BIO-187.
\bibitem{NB1}	I. Rish, An empirical study of the naive Bayes classifier, IJCAI 2001 Work. Empir. methods Artif. Intell., (2001), 41-46.
\bibitem{NB2}	J.N. Sulzmann, J. Fürnkranz, E. Hüllermeier, On pairwise naive bayes classifiers, Lect. Notes Comput. Sci. (Including Subser. Lect. Notes Artif. Intell. Lect. Notes Bioinformatics). 4701 LNAI (2007) 371–381. https://doi.org/10.1007/978-3-540-74958-5\_35.
\bibitem{GNB}	C. Bustamante, L. Garrido, R. Soto, Comparing fuzzy Naive Bayes and Gaussian Naive Bayes for decision making in RoboCup 3D, Lect. Notes Comput. Sci. (Including Subser. Lect. Notes Artif. Intell. Lect. Notes Bioinformatics). 4293 LNAI (2006) 237–247. https://doi.org/10.1007/11925231\_23.
\bibitem{MNB}	A.M. Kibriya, E. Frank, B. Pfahringer, G. Holmes, Multinomial naive bayes for text categorization revisited, Lect. Notes Artif. Intell. (Subseries Lect. Notes Comput. Sci. 3339 (2004) 488–499.
\bibitem{CNB}	J.D.M. Rennie, L. Shih, J. Teevan, D.R. Karger
Tackling the poor assumptions of Naive Bayes text classifiers, Proc. Twent. Int. Conf. Mach. Learn. ICML (2003), 616-623.
\bibitem{BNB}	V. Narayanan, I. Arora, and A. Bhatia, Fast and accurate sentiment classification using an enhanced naïve Bayes model, arXiv preprint arXiv:1305.6143, (2013). https://arxiv.org/abs/1305.6143.
\bibitem{DT}	S. Rasoul, L. David, A Survey of Decision Tree Classifier Methodology, IEEE Trans. Syst. Man. Cybern. 21 (1991) 660–674.
\bibitem{n3} D.M. Manias, M. Jammal, H. Hawilo, A. Shami, P. Heidari, A. Larabi, R. Brunner, Machine Learning for Performance-aware Virtual Network Function Placement, 2019 IEEE Glob. Commun. Conf. GLOBECOM 2019 - Proc. (2019) 12–17. https://doi.org/10.1109/GLOBECOM38437.2019.9013246.
\bibitem{IDSme}	L. Yang, A. Moubayed, I. Hamieh, A. Shami, Tree-based intelligent intrusion detection system in internet of vehicles, 2019 IEEE Glob. Commun. Conf. GLOBECOM 2019 - Proc. (2019). https://doi.org/10.1109/GLOBECOM38437.2019.9013892.
\bibitem{DTHPsk} S. Sanders, C. Giraud-Carrier, Informing the use of hyperparameter optimization through metalearning, Proc. - IEEE Int. Conf. Data Mining, ICDM. 2017-Novem (2017) 1051–1056. https://doi.org/10.1109/ICDM.2017.137.
\bibitem{RF}	M. Injadat, F. Salo, A.B. Nassif, A. Essex, A. Shami, Bayesian Optimization with Machine Learning Algorithms Towards Anomaly Detection, 2018 IEEE Glob. Commun. Conf. (2018) 1–6. https://doi.org/10.1109/glocom.2018.8647714.
\bibitem{RFour} F. Salo, M.N. Injadat, A. Moubayed, A.B. Nassif, A. Essex, Clustering Enabled Classification using Ensemble Feature Selection for Intrusion Detection, 2019 Int. Conf. Comput. Netw. Commun. ICNC 2019. (2019) 276–281. https://doi.org/10.1109/ICCNC.2019.8685636.
\bibitem{ET}	K. Arjunan, C.N. Modi, An enhanced intrusion detection framework for securing network layer of cloud computing, ISEA Asia Secur. Priv. Conf. 2017, ISEASP 2017. (2017) 1–10. https://doi.org/10.1109/ISEASP.2017.7976988.
\bibitem{XGHP} Y. Xia, C. Liu, Y.Y. Li, N. Liu, A boosted decision tree approach using Bayesian hyper-parameter optimization for credit scoring, Expert Syst. Appl. 78 (2017) 225–241. https://doi.org/10.1016/j.eswa.2017.02.017.
\bibitem{Ensemble}	T. G. Dietterich, Ensemble methods in machine learning, Mult. Classif. Syst., 1857 (2000), 1-15.
\bibitem{bagging}	A. Moubayed, E. Aqeeli, A. Shami, Ensemble-based Feature Selection and Classification Model for DNS Typo-squatting Detection, in: 2020 IEEE Can. Conf. Electr. Comput. Eng., 2020.
\bibitem{DL1}	W. Yin, K. Kann, M. Yu, and H. Schütze, Comparative Study of CNN and RNN for Natural Language Processing, arXiv preprint arXiv:1702.01923, (2017). https://arxiv.org/abs1702.01923
\bibitem{DL2}	A. Koutsoukas, K.J. Monaghan, X. Li, J. Huan, Deep-learning: Investigating deep neural networks hyper-parameters and comparison of performance to shallow methods for modeling bioactivity data, J. Cheminform. 9 (2017) 1–13. https://doi.org/10.1186/s13321-017-0226-y.
\bibitem{DL3}	T. Domhan, J.T. Springenberg, F. Hutter, Speeding up automatic hyperparameter optimization of deep neural networks by extrapolation of learning curves, IJCAI Int. Jt. Conf. Artif. Intell. 2015-January (2015) 3460–3468.
\bibitem{DL4}	Y. Ozaki, M. Yano, M. Onishi, Effective hyperparameter optimization using Nelder-Mead method in deep learning, IPSJ Trans. Comput. Vis. Appl. 9 (2017). https://doi.org/10.1186/s41074-017-0030-7.
\bibitem{DL5}	F.C. Soon, H.Y. Khaw, J.H. Chuah, J. Kanesan, Hyper-parameters optimisation of deep CNN architecture for vehicle logo recognition, IET Intell. Transp. Syst. 12 (2018) 939–946. https://doi.org/10.1049/iet-its.2018.5127.
\bibitem{TL}	D. Han, Q. Liu, W. Fan, A new image classification method using CNN transfer learning and web data augmentation, Expert Syst. Appl. 95 (2018) 43–56. https://doi.org/10.1016/j.eswa.2017.11.028.
\bibitem{ncluster}	C. Di Francescomarino, M. Dumas, M. Federici, C. Ghidini, F.M. Maggi, W. Rizzi, L. Simonetto, Genetic algorithms for hyperparameter optimization in predictive business process monitoring, Inf. Syst. 74 (2018) 67–83. https://doi.org/10.1016/j.is.2018.01.003.
\bibitem{kmeans2}	A. Moubayed, M. Injadat, A. Shami, H. Lutfiyya, Student Engagement Level in e-Learning Environment: Clustering Using K-means, Am. J. Distance Educ. 34 (2020) 1–20. https://doi.org/10.1080/08923647.2020.1696140.
\bibitem{EM}	T. K. Moon, The expectation-maximization algorithm, IEEE Signal Process. Mag. 13 (6) (1996) 47–60.
\bibitem{GMM}	S. Brahim-Belhouari, A. Bermak, M. Shi, P.C.H. Chan, Fast and Robust gas identification system using an integrated gas sensor technology and Gaussian mixture models, IEEE Sens. J. 5 (2005) 1433–1444. https://doi.org/10.1109/JSEN.2005.858926.
\bibitem{HC}	Z. Y., K. G., Hierarchical Clustering Algorithms for Document Dataset, Data Min. Knowl. Discov. 10 (2005) 141–168.
\bibitem{DBSCAN1}	K. Khan, S.U. Rehman, K. Aziz, S. Fong, S. Sarasvady, A. Vishwa, DBSCAN: Past, present and future, 5th Int. Conf. Appl. Digit. Inf. Web Technol. ICADIWT 2014. (2014) 232–238. https://doi.org/10.1109/ICADIWT.2014.6814687.
\bibitem{DBSCAN2}	H. Zhou, P. Wang, H. Li, Research on adaptive parameters determination in DBSCAN algorithm, J. Inf. Comput. Sci. 9 (2012) 1967–1973.
\bibitem{PCA}	J. Shlens, A Tutorial on Principal Component Analysis, arXiv preprint arXiv:1404.1100, (2014). https://arxiv.org/abs1404.1100
\bibitem{SVD}	N. Halko, P. Martinsson, J. Tropp, Finding structure with randomness: probabilistic algorithms for constructing approximate matrix decompositions, SIAM Rev. 53 (2) (2011), pp. 217-288
\bibitem{LDA1}	M. Loog, Conditional linear discriminant analysis, Proc. - Int. Conf. Pattern Recognit. 2 (2006) 387–390. https://doi.org/10.1109/ICPR.2006.402.
\bibitem{LDA2}	P. Howland, J. Wang, H. Park, Solving the small sample size problem in face recognition using generalized discriminant analysis, Pattern Recognit. 39 (2006) 277–287. https://doi.org/10.1016/j.patcog.2005.06.013.
\bibitem{BB2}	I. Ilievski, T. Akhtar, J. Feng, C.A. Shoemaker, Efficient hyperparameter optimization of deep learning algorithms using deterministic RBF surrogates, 31st AAAI Conf. Artif. Intell. AAAI 2017. (2017) 822–829.
\bibitem{grid1}	M.N. Injadat, A. Moubayed, A.B. Nassif, A. Shami, Systematic Ensemble Model Selection Approach for Educational Data Mining, Knowledge-Based Syst. 200 (2020) 105992. https://doi.org/10.1016/j.knosys.2020.105992.
\bibitem{grid2}	M. Injadat, A. Moubayed, A.B. Nassif, A. Shami, Multi-split Optimized Bagging Ensemble Model Selection for Multi-class Educational Data Mining, Springer’s Appl. Intell. (2020).
\bibitem{Optunity}	M. Claesen, J. Simm, D. Popovic, Y. Moreau, and B. De Moor, Easy Hyperparameter Search Using Optunity, arXiv preprint arXiv:1412.1114, (2014). https://arxiv.org/abs1412.1114. 
\bibitem{RStime} C. Witt, Worst-case and average-case approximations by simple randomized search heuristics, in: Proceedings of the 22nd Annual Symposium on Theoretical Aspects of Computer Science, STACS’05, Stuttgart, Germany, 2005, pp. 44–56.
\bibitem{GBO}	Y. Bengio, Gradient-based optimization of hyperparameters, Neural Comput. 12 (8) (2000) 1889-1900.
\bibitem{GBOtime}	H. H. Yang and S. I. Amari, Complexity Issues in Natural Gradient Descent Method for Training Multilayer Perceptrons, Neural Comput. 10 (8) (1998) 2137–2157.
\bibitem{BO1}	J. Snoek, H. Larochelle, R. Adams
Practical Bayesian optimization of machine learning algorithms
Adv. Neural Inf. Process. Syst. 4 (2012), 2951-2959.
\bibitem{BO2}	E. Hazan, A. Klivans, and Y. Yuan, Hyperparameter optimization: a spectral approach, arXiv preprint arXiv:1706.00764, (2017). https://arxiv.org/abs1706.00764.
\bibitem{GP}	M. Seeger, Gaussian processes for machine learning, Int. J. Neural Syst., 14 (2004), 69-106.
\bibitem{SMAC}	 F. Hutter, H. H. Hoos, and K. Leyton-Brown, Sequential model-based optimization for general algorithm configuration, Proc. LION 5, (2011) 507-523.
\bibitem{BOs}	I. Dewancker, M. McCourt, S. Clark, Bayesian Optimization Primer, (2015) URL: https://sigopt.com/static/pdf/SigOpt
Bayesian Optimization Primer.pdf
\bibitem{BOGPtime}	J. Hensman, N. Fusi, and N. D. Lawrence, Gaussian processes for big data, arXiv preprint arXiv:1309.6835, (2013). https://arxiv.org/abs/1309.6835.
\bibitem{HPS}	M. Claesen and B. De Moor, Hyperparameter Search in Machine Learning, arXiv preprint arXiv:1502.02127, (2015). https://arxiv.org/abs1502.02127.
\bibitem{subset}	L. Bottou, Large-scale machine learning with stochastic gradient descent, Proceedings of the COMPSTAT, Springer (2010) 177-186.
\bibitem{multifidelity} S. Zhang, J. Xu, E. Huang, C.H. Chen, A new optimal sampling rule for multi-fidelity optimization via ordinal transformation, IEEE Int. Conf. Autom. Sci. Eng. 2016-Novem (2016) 670–674. https://doi.org/10.1109/COASE.2016.7743467.
\bibitem{SH}	Z. Karnin, T. Koren, O. Somekh, Almost optimal exploration in multi-armed bandits, 30th Int. Conf. Mach. Learn. ICML 2013. 28 (2013) 2275–2283.
\bibitem{BOHB}	S. Falkner, A. Klein, F. Hutter, BOHB: Robust and Efficient Hyperparameter Optimization at Scale, 35th Int. Conf. Mach. Learn. ICML 2018. 4 (2018) 2323–2341.
\bibitem{Metaheuristic}	 A. Gogna, A. Tayal, Metaheuristics: Review and application, J. Exp. Theor. Artif. Intell. 25 (2013) 503–526. https://doi.org/10.1080/0952813X.2013.782347.
\bibitem{GA3}	F. Itano, M.A. De Abreu De Sousa, E. Del-Moral-Hernandez, Extending MLP ANN hyper-parameters Optimization by using Genetic Algorithm, Proc. Int. Jt. Conf. Neural Networks. 2018-July (2018) 1–8. https://doi.org/10.1109/IJCNN.2018.8489520.
\bibitem{goodini} B. Kazimipour, X. Li, A.K. Qin, A Review of Population Initialization Techniques for Evolutionary Algorithms, 2014 IEEE Congr. Evol. Comput. (2014) 2585–2592. https://doi.org/10.1109/CEC.2014.6900618.
\bibitem{ini1} S. Rahnamayan, H.R. Tizhoosh, M.M.A. Salama, A novel population initialization method for accelerating evolutionary algorithms, Comput. Math. with Appl. 53 (2007) 1605–1614. https://doi.org/10.1016/j.camwa.2006.07.013.
\bibitem{GAtime}	F. G. Lobo, D. E. Goldberg, and M. Pelikan, Time complexity of genetic algorithms on exponentially scaled problems, Proc. Genet. Evol. Comput. Conf., (2000) 151-158.
\bibitem{PSO1}	Y. Shi, R.C. Eberhart, Parameter Selection in Particle Swarm Optimization, Evolutionary Programming VII, Springer (1998) 591-600.
\bibitem{PSOtime}	X. Yan, F. He, Y. Chen, A Novel Hardware / Software Partitioning Method Based on Position Disturbed Particle Swarm Optimization with Invasive Weed Optimization, 32 (2017) 340–355. https://doi.org/10.1007/s11390-017-1714-2.
\bibitem{PSOdiscrete} M.Y. Cheng, K.Y. Huang, M. Hutomo, Multiobjective Dynamic-Guiding PSO for Optimizing Work Shift Schedules, J. Constr. Eng. Manag. 144 (2018) 1–7. https://doi.org/10.1061/(ASCE)CO.1943-7862.0001548.
\bibitem{ini2} H. Wang, Z. Wu, J. Wang, X. Dong, S. Yu, G. Chen, A new population initialization method based on space transformation search, 5th Int. Conf. Nat. Comput. ICNC 2009. 5 (2009) 332–336. https://doi.org/10.1109/ICNC.2009.371.
\bibitem{HBBO}	J. Wang, J. Xu, and X. Wang, Combination of Hyperband and Bayesian Optimization for Hyperparameter Optimization in Deep Learning, arXiv preprint arXiv:1801.01596, (2018). https://arxiv.org/abs1801.01596.
\bibitem{PSO3}	P. Cazzaniga, M.S. Nobile, D. Besozzi, The impact of particles initialization in PSO: Parameter estimation as a case in point, 2015 IEEE Conf. Comput. Intell. Bioinforma. Comput. Biol. CIBCB 2015. (2015) 1–8. https://doi.org/10.1109/CIBCB.2015.7300288.
\bibitem{BayesOpt}	R. Martinez-Cantin, BayesOpt: A Bayesian optimization library for nonlinear optimization, experimental design and bandits, J. Mach. Learn. Res. 15 (2015) 3735–3739.
\bibitem{Hyperopt}	J. Bergstra, B. Komer, C. Eliasmith, D. Yamins, D.D. Cox, Hyperopt: A Python library for model selection and hyperparameter optimization, Comput. Sci. Discov. 8 (2015). https://doi.org/10.1088/1749-4699/8/1/014008.
\bibitem{hyperopt-sklearn}	B. Komer, J. Bergstra, and C. Eliasmith, Hyperopt-sklearn: Automatic hyperparameter configuration for scikit-learn, Proc. ICML Workshop AutoML, (2014) 34–40.
\bibitem{hyperas} M. Pumperla, Hyperas, 2019. http://maxpumperla.com/hyperas/.
\bibitem{SMAC3} M. Lindauer, K. Eggensperger, M. Feurer, S. Falkner, A. Biedenkapp, and F. Hutter, Smac v3: Algorithm configuration in python, 2017. https://github.com/automl/SMAC3. 
\bibitem{SKOPT}	Tim Head, MechCoder, Gilles Louppe, \textit{et al.}, scikitoptimize/scikit-optimize: v0.5.2, 2018. https://doi.org/10.5281/zenodo.1207017.
\bibitem{GPflowOpt}	N. Knudde, J. van der Herten, T. Dhaene, and I. Couckuyt, GPflowOpt: A Bayesian Optimization Library using TensorFlow, arXiv preprint arXiv:1711.03845, (2017). https://arxiv.org/abs1711.03845.
\bibitem{talos} Autonomio Talos [Computer software], 2019. http://github.com/autonomio/talos.
\bibitem{SHERPA}	L. Hertel, P. Sadowski, J. Collado, P. Baldi, Sherpa: Hyperparameter Optimization for Machine Learning Models, Conf. Neural Inf. Process. Syst. (2018).
\bibitem{tf} M. Abadi, A. Agarwal, P. Barham, E. Brevdo, Z. Chen, C. Citro, \textit{et al.}, TensorFlow: Large-Scale Machine Learning on Heterogeneous Distributed Systems, arXiv preprint arXiv:1603.04467, (2016). https://arxiv.org/abs1603.04467.
\bibitem{Osprey}	J. Grandgirard, D. Poinsot, L. Krespi, J.P. Nénon, A.M. Cortesero, Osprey: Hyperparameter Optimization for Machine Learning, 103 (2002) 239–248. https://doi.org/10.21105/joss.00034.
\bibitem{FAR-HO}	L. Franceschi, M. Donini, P. Frasconi, and M. Pontil, Forward and reverse gradient-based hyperparameter optimization, 34th Int. Conf. Mach. Learn. ICML 2017, 70 (2017) 1165-1173.
\bibitem{DEAP}	F.A. Fortin, F.M. De Rainville, M.A. Gardner, M. Parizeau, C. Gagńe, DEAP: Evolutionary algorithms made easy, J. Mach. Learn. Res. 13 (2012) 2171–2175.
\bibitem{TPOT}	R. S. Olson and J. H. Moore, TPOT: A tree-based pipeline optimization tool for automating machine learning, Auto Mach. Learn. (2019) 151-160. https://doi.org/10.1007/978-3-030-05318-5\_8 
\bibitem{nevergrad} J. Rapin and O. Teytaud, Nevergrad - A gradient-free optimization platform, 2018. https://GitHub.com/FacebookResearch/Nevergrad.
\bibitem{github} L. Yang and A. Shami, Hyperparameter Optimization of Machine Learning Algorithms, 2020. https://github.com/LiYangHart/Hyperparameter-Optimization-of-Machine-Learning-Algorithms.
\bibitem{NNetime} C.M. Bishop, Neural Networks for Pattern Recognition, Oxford University Press (1995).
\bibitem{Imagenet}	A. Krizhevsky, I. Sutskever, G.E. Hinton, Imagenet classification with deep convolutional neural networks, Adv. Neural Inf. Process. Syst. 25 (2012) 1097-1105
\bibitem{COCO}	N. Hansen, A. Auger, O. Mersmann, T. Tusar, and D. Brockhoff, COCO: A Platform for Comparing Continuous Optimizers in a Black-Box Setting, arXiv preprint arXiv:1603.08785, (2016). https://arxiv.org/abs1603.08785.
\bibitem{over-fitting}	G.C. Cawley, N.L.C. Talbot, On over-fitting in model selection and subsequent selection bias in performance evaluation, J. Mach. Learn. Res. 11 (2010) 2079–2107.
\bibitem{SystemML}	M. Boehm, A. Surve, S. Tatikonda, \textit{et al.}, SystemML: declarative machine learning on spark, Proc. VLDB Endow. 9 (2016) 1425–1436. https://doi.org/10.14778/3007263.3007279.
\bibitem{MLib}	X. Meng, J. Bradley, B. Yavuz, \textit{et al.}, Mllib: machine learning in apache spark, J. Mach. Learn. Res. 17 (1) (2016) 1235-1241.

\end{thebibliography}
\end{document}